\documentclass[]{seed_like}

\usepackage{graphicx}
\usepackage{booktabs}      
\usepackage{multirow}      
\usepackage{multicol}      
\usepackage{arydshln}      
\usepackage{adjustbox}     
\usepackage{makecell}      
\usepackage{colortbl}      
\usepackage{orcidlink}
\usepackage{amsmath, amsfonts, amssymb}
\usepackage{url}
\usepackage{booktabs, multirow, arydshln, adjustbox, makecell}
\usepackage{pifont, microtype, nicefrac, xcolor}
\usepackage{enumitem, wrapfig, algorithm, algpseudocode}
\usepackage{nccmath}
\usepackage{xspace}
\usepackage{eccvabbrv}

\definecolor{lightblue}{rgb}{0.796, 0.894, 0.9808}
\definecolor{naturegreen}{RGB}{0,102,85}
\definecolor{cvprblue}{rgb}{0.21,0.49,0.74}
\definecolor{lightblue}{rgb}{0.796, 0.894, 0.9808}
\definecolor{forestgreen}{RGB}{34,139,34}
\definecolor{crimson}{RGB}{220,20,60}
\definecolor{wine}{HTML}{830E0D}

\newcommand{\tablebf}[1]{{\normalfont\bfseries #1}}
\newcommand{\best}[1]{\tablebf{#1}}
\newcommand{\second}[1]{\underline{#1}}
\newcommand{\third}[1]{\textit{#1}}

\title{Layer-Specific Prompt Fusion Discovery via Differentiable Search in Vision Foundation Models}

\author[1]{Xi Xiao}
\author[2]{Xingjian Li}
\author[3]{Yunbei Zhang}
\author[4]{Cheng Han}
\author[5]{Tianming Liu}
\author[1\dag]{\protect\\ Tianyang Wang}
\author[2]{Runmin Jiang}
\author[3]{Jihun Hamm}
\author[6]{Xiao Wang}
\author[2\dag]{Min Xu}

\affiliation[1]{UAB}
\affiliation[2]{CMU}
\affiliation[3]{Tulane}
\affiliation[4]{UMKC}
\affiliation[5]{UGA}
\affiliation[6]{ORNL}

\contribution[\dag]{Co-advising}

\abstract{
Visual prompt tuning has emerged as a parameter-efficient fine-tuning approach for adapting large-scale Vision Transformers (ViTs) to downstream tasks. As its learnable prompts are applied in input and feature spaces, prior to jointly going through attention in transformer layers, the most commonly used scheme for fusing image and prompt tokens is concatenation or addition. In this paper, we aim to study a fundamental yet essential problem in visual prompt tuning: whether a single fusion scheme tends to yield better results, and whether that would be beneficial to develop a hybrid fusion scheme. To this end, we formulate the task as a bi-level optimization problem, and solve it leveraging differentiable architecture search. In this context, the learnable prompts and their fusion schemes are jointly optimized. To enrich the search space in the architecture search, we propose two additional fusion schemes, namely, affine transformation and cross-attention, in addition to concatenation and addition. Extensive experiments on 34 datasets spanning VTAB-1k, FGVC, and HTA show consistent gains over prompt-tuning baselines. With a frozen ViT backbone, our method delivers a favorable accuracy--latency--parameter trade-off compared with VPT-Deep and recent variants. Our findings reveal that how prompts fuse with image tokens plays a significant role in visual prompt tuning, and a hybrid fusion fashion can more effectively leverage layer semantics of ViTs, contributing a novel perspective for visual prompt-tuning research.

}

\date{\today}
\correspondence{Xi Xiao (\email{xxiao@uab.edu})}

\checkdata[Project Page]{\url{https://xixiaouab.github.io/Prompt-Fusion-Discovery/}}

\begin{document}
\maketitle

\makeatletter
\let\old@vspace\@vspace
\let\old@vspacer\@vspacer
\renewcommand{\@vspace}[1]{}
\renewcommand{\@vspacer}[1]{}
\makeatother

\section{Introduction}
\label{sec:intro}

Vision transformers (ViTs) have demonstrated impressive performance across a wide range of visual recognition tasks~\cite{dosovitskiyimage, liu2021swin, wangvisual}. However, if full fine-tuning is employed, adapting such models to downstream scenarios inevitably demands high computing load, and a potential overfitting problem
to small downstream datasets. \emph{Visual Prompt Tuning}~(VPT)~\cite{jia2022visual}, a \textit{Parameter-efficient fine-tuning}~(PEFT) solution, has recently emerged to address this challenge.
VPT incorporates learnable prompts (\ie, a certain type of parameters) into input and feature spaces, and solely updates these prompts during fine-tuning. The core operation in VPT is how to fuse the learnable prompts with image tokens and/or feature tokens before the fused elements are fed into transformer layers. To date, the most commonly used scheme is either concatenation or addition \cite{jia2022visual, liu2026all, xiao2026focus, xiao2026self}.

Nevertheless, it remains unknown whether these fusion schemes contribute differently to ViT's fine-tuning. If so, should we always use a single fusion scheme or is there an optimal hybrid fusion fashion that takes the best of both concatenation and addition to fully leverage \textbf{layer-wise} information \cite{raghu2021vision, skean2025layer, liu2026dispersion}? Moreover, if a hybrid fusion scheme can be optimal, what else single fusion scheme can be developed to facilitate the design of the hybrid fashion? Such questions are fundamental yet essential to visual prompt tuning research, which are rarely explored in previous literature.


In this work, we aim to answer the aforementioned questions in a unified task, and formulate the task as a bi-level optimization problem, consisting of inner and outer optimizations. In the former, the learnable prompts will be optimized, while in the latter, the best combination of fusion schemes will be learned. We tackle this problem by developing a novel method with differentiable neural architecture search (NAS), fully leveraging layer-wise information, where another challenge naturally arises:
how to design a search space that is indispensable in NAS? A natural solution is to incorporate the two commonly used single fusion schemes directly.
However, combining these two simple fusions is unlikely to yield optimal results (Sec.~\ref{sec:ablation}).
To enrich the search space to discover better hybrid prompting fashions, we thus propose two new single fusion schemes, namely, affine transformation and cross-attention. The rationale lies in that they are simple and light-weight, and can be seamlessly used to fuse prompt and image tokens efficiently in the VPT context (Sec.~\ref{sec:efficiency_latency}). 
Together, our proposed method is highly scalable and can accommodate any new single fusion schemes by including them in the search space.



We validate the proposed method from both interpretive and empirical perspectives. We interpret our method through the lens of \textit{Information Bottleneck} (IB)~\cite{tishby2000information}, and observe that jointly optimizing prompts and fusion schemes 
leads to a lower empirical IB surrogate 
than optimizing prompts alone, supporting the role of hybrid fusion in prompt tuning. Empirically, we conduct extensive experiments on 34 datasets across three widely used benchmarks: VTAB-1k~\cite{zhai2019large}, FGVC~\cite{wah2011caltech}, and HTA~\cite{huang2023diversity}. The results (\eg, 77.01\% on VTAB-1k mean, 92.5\% on HTA mean, 91.6\% on FGVC mean, and only 0.75\% of the ViT's parameters tuned) reveal that our method consistently and significantly outperforms existing state-of-the-art baselines. More importantly, these results help answer the aforementioned crucial questions as follows: \textbf{\ding{172}} there is an optimal hybrid fusion fashion that is significantly better than single fusion scheme in visual prompt tuning; and \textbf{\ding{173}} the hybrid fashion is highly scalable and can seamlessly include new single fusion schemes. 


Our main contributions are summarized as follows:
\begin{itemize}
    \item We raise fundamental yet essential questions on fusing prompt and image tokens in visual prompt tuning research, and reveal that using a single fusion scheme fails to 
    yield optimal performance.
    \item We propose a hybrid fusion method and formulate it as a bi-level optimization problem. To solve the new optimization problem, we develop a novel method with differentiable neural architecture search.
    \item We provide both a theoretical interpretation (via Information Bottleneck) and extensive empirical evaluations to demonstrate our method's state-of-the-art and parameter-efficient performance across diverse datasets. We use the observed results, along with the analysis, to answer the raised key questions in visual prompt tuning research.
\end{itemize}

\section{Method}
\label{sec:method}

\begin{figure*}[t]
    \centering
    \includegraphics[width=0.98\linewidth]{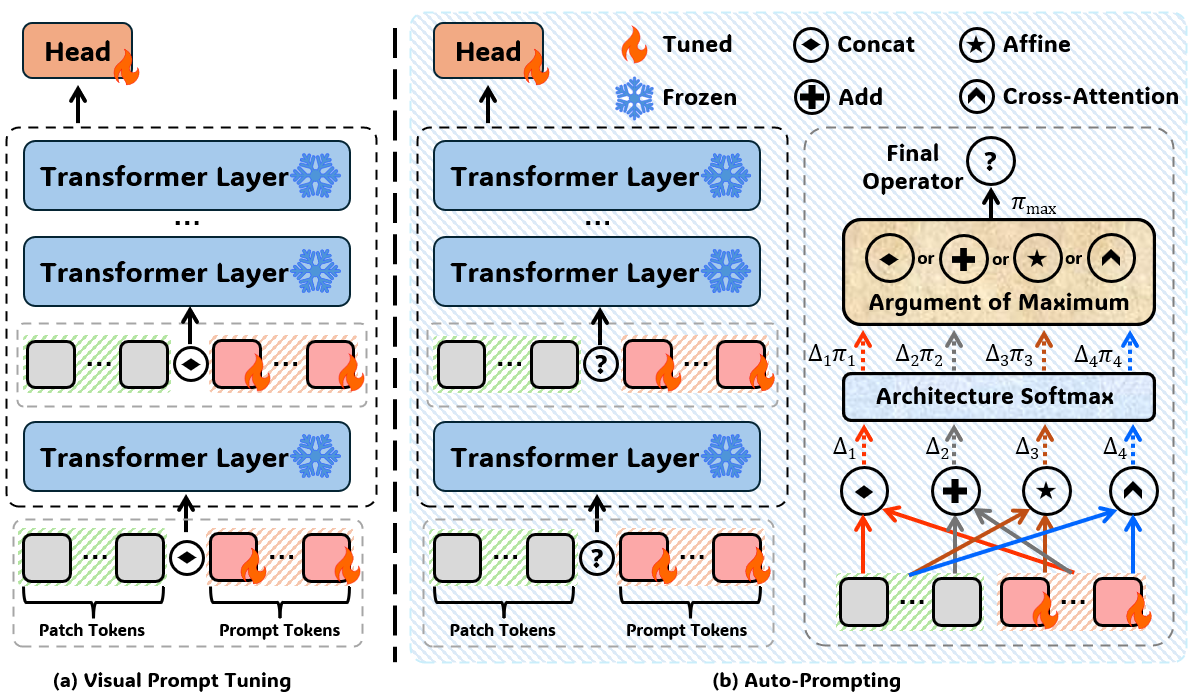}
    \caption{Overview of standard Visual Prompt Tuning \textit{$vs.$} Auto-Prompting framework. \textbf{(a) Visual Prompt Tuning:} Employs a \textit{single, fixed fusion operator} to combine learnable prompt tokens with patch tokens at every layer. The Transformer Layers remain frozen. \textbf{(b) Auto-Prompting (Ours):} We replace the fixed operator with a \textit{searchable fusion module} that operates in two phases. \textbf{Search Phase:} An `Architecture Softmax' module generates layer-wise preference weights ($\pi_1, \pi_2, \pi_3, \pi_4$) for a set of fundamental operators (Concat $\diamond$, Add $+$, Affine $\star$, Cross-Attn $\wedge$). Concurrently, the operators process the inputs in parallel to produce their respective outputs ($\Delta_1, \Delta_2, \Delta_3, \Delta_4$). \textbf{Discretization Phase:} After training, we discretize each layer by selecting the final operator with the largest learned preference.}
    \label{fig:darts_pt_architecture}
\end{figure*}

We propose a generalized prompt tuning framework that \emph{treats the fusion between prompts and visual tokens as a searchable, differentiable component} (Figure~\ref{fig:darts_pt_architecture}). Concretely, each Transformer layer chooses one of several candidate fusion operators that determine \emph{how} prompts interact with tokens before entering the frozen ViT block.
This section clarifies the injection point and notation, specifies the operator set (Sec.~\ref{sec:ops}), introduces Differentiable Architecture Search (DARTS)~\cite{liudarts} and our bilevel optimization with stability regularizers (Sec.~\ref{sec:darts}), and gives an \emph{Information Bottleneck} (IB) perspective (Sec.~\ref{sec:ib}) on why learning the fusion rule improves adaptation.

Let $f_{\psi}$ be a frozen ViT encoder.
At block $l$, let $x^{(l-1)}\!\in\!\mathbb{R}^{k\times d}$ be the token sequence (incl.\ CLS) and
$p^{(l)}\!\in\!\mathbb{R}^{m\times d}$ be $m$ learnable prompt tokens.
We use the Pre-LayerNorm (Pre-LN) form, where a standard block is:
\begin{equation}
\begin{aligned}
y        &= x^{(l-1)} + \mathrm{MSA}\!\big(\mathrm{LN}(x^{(l-1)})\big),\\
x^{(l)}  &= y          + \mathrm{MLP}\!\big(\mathrm{LN}(y)\big).
\end{aligned}
\end{equation}

Denote $\tilde{x}^{(l-1)}=\mathrm{LN}(x^{(l-1)})$.
We inject fusion \emph{after} this internal LayerNorm and \emph{before} the frozen MSA sublayer:
\begin{equation}
\begin{aligned}
\widehat{x}_{\mathrm{msa}}^{(l-1)} &= \Delta^{(l)}\!\big(p^{(l)},\,\tilde{x}^{(l-1)}\big),\\
y                                   &= x^{(l-1)} + \mathrm{MSA}\!\big(\widehat{x}_{\mathrm{msa}}^{(l-1)}\big).
\end{aligned}
\end{equation}

No extra normalization is applied after $\Delta^{(l)}$.
By default, we inject fusion once per block (before MSA). We split trainable variables into two groups.
(1) \emph{Model parameters} $\phi$: prompt tokens $\{p^{(l)}\}$ and operator-internal weights
(\eg, the reduction matrix $R^{(l)}$ in \texttt{concat}, FiLM MLPs in \texttt{affine}, and
$W_Q,W_K,W_V$ in \texttt{cross} when not shared).
(2) \emph{Architecture parameters} $\alpha=\{\alpha^{(l)}\}_{l=1}^{L}$: for each layer $l$,
$\alpha^{(l)}\!\in\!\mathbb{R}^{|\mathcal S|}$ contains one logit (unnormalized preference) per
candidate operator in $\mathcal S=\{\texttt{concat},\texttt{add},\texttt{affine},\texttt{cross}\}$.
These logits are converted into mixing weights by:
\[
\pi^{(l)} = \mathrm{softmax}\!\big(\alpha^{(l)}/\tau\big), \ \ \tau>0,\quad
\pi^{(l)}_i \ge 0,\ \sum_{i\in\mathcal S}\pi^{(l)}_i = 1,
\]
and define the soft fusion at layer $l$:
\[
\Delta^{(l)}_{\text{soft}}(p^{(l)},\,\tilde{x}^{(l-1)})
= \sum_{i\in\mathcal S}\pi^{(l)}_i\,\Delta_i\!\big(p^{(l)},\,\tilde{x}^{(l-1)}\big).
\]

At discretization we pick $\widehat{i}^{(l)}=\operatorname*{arg\,max}_{i\in\mathcal S}\pi^{(l)}_i$ and keep only $\Delta_{\widehat{i}^{(l)}}$.

\subsection{Fusion Operator Design}
\label{sec:ops}

We place a fusion module $\Delta^{(l)}(\cdot)$ before each frozen block, with the constraint that its output is always $k\times d$ (token count $k$ incl.\ CLS) so the backbone interface never changes. We consider
\[
\mathcal{S}=\{\Delta_{\texttt{concat}},\,\Delta_{\texttt{add}},\,\Delta_{\texttt{affine}},\,\Delta_{\texttt{cross}}\}.
\]

This set covers four complementary fusion operators while keeping the interface fixed:
\texttt{concat} augments the sequence and mixes prompt cues with a light token-preserving reduction;
\texttt{add} injects a prompt-dependent residual bias at negligible cost;
\texttt{affine} applies prompt-conditioned channel-wise scale–shift (with \texttt{add} as a special case);
and \texttt{cross-attention} enables content-adaptive routing from a prompt memory when stronger fusion is needed.
The motivation for involving these four operators is that many heavier fusion designs (\eg, gated residuals, dynamic projections) can be \textbf{precisely composed} from or \textbf{closely approximated} by these four primitives.
Searching within this basis lets each layer select the right fusion without redesigning the frozen ViT and exposes an accuracy–cost frontier that we regularize during search (Sec.~\ref{sec:darts}). We elaborate on them in detail below.
\paragraph{(1) \texttt{concat}.}
Unlike the original VPT that feeds $(m{+}k)$ tokens into the frozen block, we keep the backbone interface fixed at $k\times d$ to avoid changing positional embeddings, attention map size, and FLOPs inside the block. We first prefix prompts and then reduce back to $k$ tokens with a light, token preserving mixer:
\begin{subequations}\label{eq:concat}
\begin{gather}
\widetilde{x} = [\,p^{(l)};\,\tilde{x}^{(l-1)}\,],\\
R^{(l)} = \operatorname{softmax}_{\mathrm{col}}\!\big(U^{(l)}\big),\\
\Delta_{\texttt{concat}}\!\big(p^{(l)},\tilde{x}^{(l-1)}\big) = (R^{(l)})^{\!\top}\,\widetilde{x}.
\end{gather}
\end{subequations}

Here $U^{(l)}\!\in\!\mathbb{R}^{(m+k)\times k}$ collects the per-column logits and
$R^{(l)}=\operatorname{softmax}_{\mathrm{col}}(U^{(l)})$ is column-stochastic (each column is nonnegative and sums to $1$),
so every output token is a convex combination of the $(m{+}k)$ inputs. This operator uses no $W_Q/W_K/W_V$ projections and is therefore cheaper and more stable than full cross-attention. By default $R^{(l)}$ is layer-shared to keep parameters minimal (More analysis in Appendix).

\paragraph{(2) \texttt{add}.}
We summarize prompts by $s=\mathrm{LN}(\mathrm{mean}(p^{(l)}))\in\mathbb{R}^{d}$ and broadcast it across tokens:
\begin{equation}
\Delta_{\texttt{add}}(p^{(l)},\tilde{x}^{(l-1)}) \;=\; \tilde{x}^{(l-1)} \;+\; \mathbf{1}_{k}\, s^\top ,
\label{eq:add}
\end{equation}
where $\mathbf{1}_{k}$ repeats $s$ along the token axis.

\paragraph{(3) \texttt{affine}.}
Feature-wise Linear Modulation (FiLM) applies channel-wise scale and shift conditioned on an external signal. We reuse the same prompt summary $s$ as in Eq.~\eqref{eq:add} and generate per-channel scale/bias via two small MLPs:
\begin{equation}
\phi_{t}(s)=W^{2}_{t}\,\sigma_h\!\big(W^{1}_{t}s+b^{1}_{t}\big)+b^{2}_{t},
\quad t\in\{\gamma,\beta\}.
\end{equation}

Then
\begin{ceqn}
\begin{equation}
\label{eq:affine}
\begin{aligned}
\gamma &= \sigma\!\big(\phi_\gamma(s)\big),\quad \beta=\phi_\beta(s),\\
\Delta_{\texttt{affine}}\!\big(p^{(l)},\tilde{x}^{(l-1)}\big)
&= \gamma \odot \tilde{x}^{(l-1)} + \mathbf{1}_{k}\,\beta^\top ,
\end{aligned}
\end{equation}
\end{ceqn}
with $\sigma$ a squashing nonlinearity and $\odot$ the Hadamard product. Eq.~\eqref{eq:add} is the special case $\gamma=\mathbf{1}$ and $\beta=s$.
The summary vector is used only for \texttt{add} and \texttt{affine}, whose outputs are channel-wise modulations over the fixed $k\times d$ image-token interface. In contrast, \texttt{concat} and \texttt{cross-attention} consume all $m$ prompt tokens directly. The LayerNorm-mean summary gives a permutation-invariant task descriptor and prevents scale drift in these modulation branches; Appendix~\ref{app:ops} reports ablations against first-token, unnormalized mean, and raw-token conditioning.

\paragraph{(4) \texttt{cross-attention} (prompt-as-memory).}
With $H$ heads and $d_h{=}d/H$, image tokens attend to a prompt memory per head:
\begin{equation}\label{eq:cross}
\begin{aligned}
Q^{(h)}_x &= \tilde{x}^{(l-1)} W^{(h)}_Q \in \mathbb{R}^{k\times d_h},\\
K^{(h)}_p &= p^{(l)}           W^{(h)}_K \in \mathbb{R}^{m\times d_h},\\
V^{(h)}_p &= p^{(l)}           W^{(h)}_V \in \mathbb{R}^{m\times d_h},\\
A^{(h)}   &= \mathrm{softmax}\!\Big(\tfrac{Q^{(h)}_x (K^{(h)}_p)^\top}{\sqrt{d_h}}\Big)\in\mathbb{R}^{k\times m},\\
O^{(h)}   &= A^{(h)} V^{(h)}_p \in \mathbb{R}^{k\times d_h}.
\end{aligned}
\end{equation}

We then concatenate head outputs along the channel dimension and add a residual:
\[
\Delta_{\texttt{cross}}\!\big(p^{(l)},\tilde{x}^{(l-1)}\big)
= \big[\mathrm{Concat}_{h=1}^H O^{(h)}\big] \;+\; \tilde{x}^{(l-1)}
\ \in\ \mathbb{R}^{k\times d}.
\]

The projection matrices $\{W^{(h)}_Q,W^{(h)}_K,W^{(h)}_V\}$ are shared across layers unless stated (More details see Appendix.~\ref{app:ops-prop}).


\subsection{Differentiable Layer-wise Operator Search}
\label{sec:darts}

Differentiable Architecture Search (DARTS)~\cite{liudarts} relaxes discrete operator choices into a softmax over learnable logits, producing mixing weights per layer; model weights are optimized on the training split and architecture parameters on the validation split, yielding a bilevel program that is later discretized by an $\arg\max$ choice. For layer $l$, architecture logits $\alpha^{(l)}\!\in\!\mathbb{R}^{|\mathcal{S}|}$ parameterize a soft mixture:
\begin{align}
\pi^{(l)}_i(\tau) &= \mathrm{softmax}\!\big(\alpha^{(l)}/\tau\big)_i, \nonumber\\
\Delta_{\!\text{soft}}^{(l)}(p^{(l)},\tilde{x}^{(l-1)};\alpha^{(l)},\tau)
&=\sum_{i\in\mathcal{S}} \pi^{(l)}_i(\tau)\,\Delta_i(p^{(l)},\tilde{x}^{(l-1)}),
\label{eq:softmix}
\end{align}
where the temperature $\tau$ is annealed from high (exploration) to low (selection). We adopt a DARTS–style bilevel objective~\cite{liudarts}:
\begin{align}
\min_{\alpha}\;&\ \mathcal{L}_{\mathrm{val}}\big(\phi^{\ast}(\alpha),\alpha\big),
\quad
\text{s.t.}\ \
\phi^{\ast}(\alpha)=\arg\min_{\phi}\mathcal{L}_{\mathrm{train}}(\phi,\alpha).
\label{eq:bilevel}
\end{align}

Gradients $\nabla_{\alpha}\mathcal{L}_{\mathrm{val}}$ use the one-step unrolled approximation with Hessian–vector \cite{boyd2004convex} products (See Appendix~\ref{app:implicit}):
\begin{align}
\phi' &= \phi - \eta_\phi \nabla_{\phi}\mathcal{L}_{\text{train}}, \\[-2pt]
\nabla_{\alpha}\mathcal{L}_{\text{val}}
&\approx
\nabla_{\alpha}\mathcal{L}_{\text{val}}(\phi',\alpha)
-\eta_\phi
\big[\nabla^2_{\alpha,\phi}\mathcal{L}_{\text{train}}\big]
\nabla_{\phi}\mathcal{L}_{\text{val}}(\phi',\alpha).
\label{eq:unroll}
\end{align}

To avoid premature collapse and to softly prefer cheaper operators when accuracies are similar, we use two unweighted terms:
\begin{subequations}\label{eq:reg}
\begin{align}
\widetilde{\mathcal R}_{\mathrm{ent}}(\alpha) &= - \sum_{l} H\!\big(\pi^{(l)}\big), \\
\widetilde{\mathcal R}_{\mathrm{cost}}(\alpha) &= \sum_{l}\sum_{i\in\mathcal S} c_i\,\pi^{(l)}_i,
\end{align}
\end{subequations}
where $H(\cdot)$ is Shannon entropy \cite{bromiley2004shannon}. We set the normalized cost prior to
$c=[0,0.06,0.30,1.00]$ for the ordered operators
$(\texttt{add},\texttt{affine},\texttt{concat},\texttt{cross})$, based on per-operator microbenchmarks.
The outer objective places the weights here:
\begin{equation}
\mathcal L_{\text{outer}}
= \mathcal L_{\mathrm{val}}
+ \lambda_{\mathrm{ent}}\,\widetilde{\mathcal R}_{\mathrm{ent}}
+ \lambda_{\mathrm{cost}}\,\widetilde{\mathcal R}_{\mathrm{cost}}.
\end{equation}

Positive $\lambda_{\mathrm{ent}}$ on the negative-entropy term encourages exploration; as the temperature $\tau$ is annealed the mixture sharpens toward a discrete choice. The cost term biases against unnecessary heavy operators when accuracies are similar. At epoch $e_\text{disc}$, we discretize each layer by $\widehat{i}^{(l)}=\arg\max_i \pi^{(l)}_i(\tau)$, freeze $\alpha$, and continue optimizing $\phi$ for a few epochs with the selected operators only. This yields a compact, deployable blueprint with the same inference path length as a fixed-fusion method (More details in Appendix~\ref{app:implicit}).

\begin{figure}[t]
  \centering
  \includegraphics[width=0.8\linewidth]{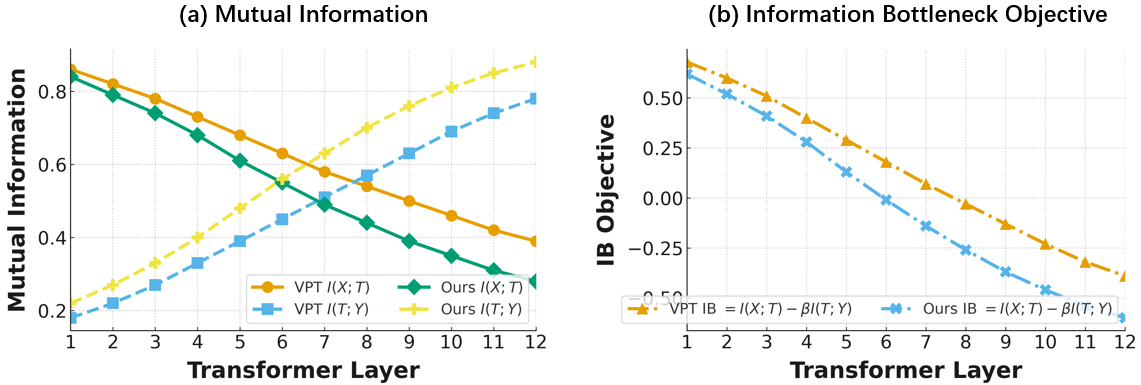}
  \caption{\textbf{Mutual information and per-layer IB objective (CUB-200, ViT-B/16).}
  Comparison between VPT \cite{jia2022visual} and our search-based fusion.
  (a) $\widehat{I}(T^{(l)};X)$ (noise compression) decreases and $\widehat{I}(T^{(l)};Y)$ (semantic retention) increases more markedly for our method in deeper blocks.
  (b) The resulting empirical IB surrogate remains consistently lower for ours, suggesting a favorable compression--relevance trade-off under a frozen backbone.}
  \label{fig:ib_curve}
\end{figure}

\subsection{Observation: From the Information Bottleneck (IB) Perspective}
\label{sec:ib}

Let $T^{(l)}=\Delta^{(l)}\!\big(p^{(l)},\tilde{x}^{(l-1)}\big)$ be the representation entering the $l$-th frozen block.
We adopt a per-layer Information Bottleneck (IB) objective \cite{tishby2000information}:
\begin{equation}
\mathcal{L}_{\mathrm{IB}}^{(l)}\!\big(T^{(l)}\big)
= I\!\big(T^{(l)};X\big)\;-\;\beta\,I\!\big(T^{(l)};Y\big),\qquad \beta>0,
\label{eq:ib}
\end{equation}
where $X$ is the input image, $Y$ the label, and $I(\cdot;\cdot)$ denotes mutual information.
Since exact MI is intractable, we report the variational surrogate
$\widehat{\mathcal{L}}_{\mathrm{IB}}^{(l)}=\widehat{I}\!\big(T^{(l)};X\big)-\beta\,\widehat{I}\!\big(T^{(l)};Y\big)$
estimated with standard MI estimators (\ie, MINE/InfoNCE/CLUB; details in Appendix~\ref{app:mi}).

\paragraph{Why learning the fusion operator under IB?}
Fix prompts $p$ and consider the family $\{T_\Delta\}$ induced by $\Delta\!\in\!\mathcal{S}$. Allowing each layer to select its own fusion operator provides a broader range of possible representations than a single fixed operator. Figure.~\ref{fig:ib_curve} reports the MI metrics on CUB-200 (ViT-B/16, $\beta{=}1$).
As shown in Fig.~\ref{fig:ib_curve}(a), our model exhibits a larger drop in $\widehat{I}\!\big(T^{(l)};X\big)$ in deep layers while maintaining higher $\widehat{I}\!\big(T^{(l)};Y\big)$. This pattern suggests stronger compression of input-dependent variation together with better label alignment. The consistently lower $\widehat{\mathcal{L}}_{\mathrm{IB}}^{(l)}$ is therefore empirical evidence that search-based fusion reaches a more favorable compression--relevance trade-off, rather than a causal proof of background removal or accuracy gains.

\paragraph{Operator schedule as information flow.}
The learned operator schedule induces a coherent information flow: shallow blocks tend to use lightweight primitives (\texttt{concat}/\texttt{add}) that preserve structural cues, whereas deeper blocks favor semantic primitives (\texttt{affine}/\texttt{cross-attention}) that emphasize label-relevant patterns. This layer-wise shaping is consistent with the IB view and with the attention trends in Sec.~\ref{sec:analysis}.

\subsection{Summary of Training Procedure.}
As detailed in Algorithm~\ref{alg:search}, our framework operates in two distinct stages.
First, the differentiable search phase jointly optimizes the prompts and the architecture logits via a bilevel objective.
Following this, the discretization step selects the dominant fusion operator for each layer, succeeded by a final short fine-tuning phase to recover performance with the pruned architecture.

\begin{algorithm}[t]
\caption{Layer-wise Search and Discretization for Prompt--Token Fusion}
\label{alg:search}
\begin{algorithmic}[1]
\Require Frozen ViT $f_\psi$; operator set $\mathcal S$; cost prior $c=\{c_i\}$; train/val splits
\Require Init prompts $\{p^{(l)}\}$ and operator-internal weights in $\phi$; architecture logits $\alpha=\{\alpha^{(l)}\}$
\Require Hyperparams: learning rates $\eta_\phi,\eta_\alpha$; temperatures $\tau_{\max},\tau_{\min}$; weights $\lambda_{\mathrm{ent}},\lambda_{\mathrm{cost}}$; search epochs $E_{\text{search}}$; fine-tune epochs $E_{\text{ft}}$
\State $\tau \leftarrow \tau_{\max}$
\Comment{\textbf{Phase I: differentiable search}}
\For{$e=1$ \textbf{to} $E_{\text{search}}$}
  \State \textbf{(Inner/train)} Compute $\pi^{(l)}=\mathrm{softmax}(\alpha^{(l)}/\tau)$ and $\Delta_{\text{soft}}^{(l)}$ (Eq.~\ref{eq:softmix}); update
  $\phi \leftarrow \phi - \eta_\phi \nabla_\phi \mathcal L_{\text{train}}(\phi,\alpha)$
  \State \textbf{(Outer/val)} Build $\widetilde{\mathcal R}_{\mathrm{ent}},\widetilde{\mathcal R}_{\mathrm{cost}}$ (Eq.~\ref{eq:reg}); compute one-step unrolled gradient (Eq.~\ref{eq:unroll}); update
  $\alpha \leftarrow \alpha - \eta_\alpha \nabla_\alpha\!\big[\mathcal L_{\mathrm{val}} + \lambda_{\mathrm{ent}}\widetilde{\mathcal R}_{\mathrm{ent}} + \lambda_{\mathrm{cost}}\widetilde{\mathcal R}_{\mathrm{cost}}\big]$
  \State Anneal $\tau \leftarrow \mathrm{CosAnneal}(\tau_{\max}\!\to\!\tau_{\min},\,e,\,E_{\text{search}})$
\EndFor
\State \textbf{Discretize}: for all $l$, set $\widehat{i}^{(l)} \leftarrow \arg\max_{i\in\mathcal S}\pi^{(l)}_i$; freeze $\alpha$ and \emph{discard} inactive operators

\State \Comment{\textbf{Phase II: short fine-tuning with discrete operators}}
\For{$e=1$ \textbf{to} $E_{\text{ft}}$}
  \State Forward with $\Delta^{(l)} \!=\! \Delta_{\widehat{i}^{(l)}}$ only; update
  $\phi \leftarrow \phi - \eta_\phi \nabla_\phi \mathcal L_{\text{train}}(\phi,\,\widehat{i}^{(1{\ldots}L)})$
\EndFor
\State \textbf{Return}: prompts $\phi^\star$ and discrete operators $\{\widehat{i}^{(l)}\}_{l=1}^{L}$
\end{algorithmic}
\end{algorithm}

\subsection{Discussion on Operator Orthogonality and Redundancy}

While the inclusion of both `Add' and `Affine' operators might suggest a functional overlap, maintaining this diversity in the search space $\mathcal{S}$ is essential for both optimization stability and computational efficiency. We justify this design from three primary perspectives. \ding{182} Regarding the optimization landscape and gradient flow, although the `Add' operator can be viewed as a special case of `Affine' transformation, optimizing `Affine' parameters ($\phi_{\gamma}, \phi_{\beta}$) via MLPs from scratch presents a significantly more complex optimization landscape. Following the principles of Residual Learning~\cite{he2016deep}, the parameter-free `Add' acts as a stable identity-like connection that facilitates smooth gradient flow during the early stages of architecture search. Furthermore, our \ding{183} outer objective leverages a cost-driven regularizer ($\mathcal{R}_{cost}$) to exploit this apparent redundancy for implicit pruning. In layers where the extra expressivity of an `Affine' calibration does not yield a significant reduction in validation loss, the optimization naturally gravitates toward the zero-cost `Add' operator. Removing the simpler `Add' would force the model to employ the heavier `Affine' operator (which incurs approximately 2$\times$ the operator latency) even for basic bias injections, leading to unnecessary computational overhead~\cite{perez2018film}. Additionally, \ding{184} empirical evidence confirms that a diverse basis set is vital for stable search dynamics, as shown in Appendix Table~\ref{tab:stability_ablation}. Without the stabilization and regularization strategies that utilize this operator redundancy, the search process suffers a 65\% collapse rate to suboptimal solutions. In summary, these four operators form a complete basis set for feature modulation: `Add' for bias injection~\cite{he2016deep}, `Affine' for feature calibration~\cite{perez2018film}, and `Cross-Attention' for dynamic content-adaptive routing~\cite{vaswani2017attention}.

\section{Experiments}
\label{sec:experiments}



\subsection{Experimental Setup}
\label{sec:setup}

\noindent\textbf{Benchmarks.}
We follow VPT~\cite{jia2022visual} on VTAB-1k~\cite{zhai2019large} (19 datasets across Natural/Specialized/Structured), and include fine grained datasets such as CUB-200-2011~\cite{wah2011caltech} and Oxford Flowers-102~\cite{nilsback2008automated}. To test robustness with different pretraining biases, we additionally evaluate under MAE~\cite{he2022masked} and MoCo v3~\cite{chen2021empirical}. For architectural generality, we further report results on a hierarchical backbone (Swin-Base~\cite{liu2021swin}).

\noindent\textbf{Fusion candidates and search.}
Each Transformer layer selects its fusion operator from \texttt{concat}, \texttt{add}, \texttt{affine}, and \texttt{cross-attention} via a differentiable relaxation (Sec.~\ref{sec:method}). The backbone stays frozen throughout. We adopt a bilevel schedule: inner loop updates optimize prompts and fusion weights on training data, while outer loop updates refine architecture logits on held out validation data. A temperature annealing is applied to stabilize discretization toward a single operator per layer at convergence.

\noindent\textbf{Training details.}
We use AdamW with learning rate $1\!\times\!10^{-3}$, weight decay $0.01$, cosine schedule, $100$ epochs, batch size $64$. We report top-1 accuracy averaged over three seeds per dataset. All runs are on NVIDIA A100 GPUs. To ensure comparability, Tuned/\% includes the task-specific head and follows the macro averaging convention of the compared works; per task counts are provided in Appendix~\ref{app:Per1}.

\begin{table*}[!htbp]
\centering
    \caption{
    \textbf{Comparison of fine-tuning methods with ViT-Base/16 backbone.} \textbf{Bold} and \underline{underlined} indicate best and second-best results. We report mean accuracy across datasets for each benchmark (FGVC: 5 datasets, HTA: 10 datasets, VTAB-1k: 19 datasets) and for each VTAB-1k category (Natural, Specialized, and Structured). See Appendix.~\ref{app:Per1} for per-task results.
    Same for Table \ref{table:mae_moco}.}
\vspace{-5pt}
\label{tab:full_comparison}

\scriptsize 

\resizebox{\textwidth}{!}{%
\begin{tabular}{c||c|c|c|c|ccc|c}
\Xhline{4\arrayrulewidth}
\rowcolor{gray!20}
ViT-Base/16~\cite{dosovitskiyimage} & Tuned/Total &  & FGVC Mean & HTA Mean & \multicolumn{3}{c|}{VTAB-1k~\cite{zhai2019large}} & VTAB-1k Mean \\
\rowcolor{gray!20}
\rowcolor{gray!20}
(85.8M) & (\%) & \multirow{-2}{*}{Extra Params} & (\%) & (\%) & \textit{Natural} & \textit{Specialized} & \textit{Structured} & (\%) \\
\hline \hline
Full \textcolor{lightgray}{\scriptsize{[CVPR22]}}\cite{iofinova2022well} & 100.00 & — & 88.54 & 85.8 & 75.88 & 83.36 & 47.64 & 65.57 \\
\hline
Linear \textcolor{lightgray}{\scriptsize{[CVPR22]}}\cite{iofinova2022well} & 0.08 & — & 79.32 & 75.7 & 68.93 & 77.16 & 26.84 & 52.94 \\
Partial-1 \textcolor{lightgray}{\scriptsize{[NeurIPS14]}}\cite{yosinski2014transferable} & 8.34 & — & 82.63 & 80.8 & 69.44 & 78.53 & 34.17 & 56.52 \\
MLP-3 \textcolor{lightgray}{\scriptsize{[CVPR20]}}\cite{chen2020improved} & 1.44 & \checkmark & 79.80 & 78.5 & 67.80 & 72.83 & 30.62 & 53.21 \\
\hline
Sidetune \textcolor{lightgray}{\scriptsize{[ECCV20]}}\cite{zhang2020side} & 10.08 & — & 78.35 & 72.3 & 58.21 & 68.12 & 23.41 & 45.65 \\
Bias \textcolor{lightgray}{\scriptsize{[NeurIPS17]}}\cite{rebuffi2017learning} & 0.80 & — & 88.41 & 82.1 & 73.30 & 78.25 & 44.09 & 62.05 \\
Adapter \textcolor{lightgray}{\scriptsize{[NeurIPS20]}}\cite{cai2020tinytl} & 1.02 & \checkmark & 85.46 & 80.6 & 70.67 & 77.80 & 33.09 & 62.41 \\
LoRA \textcolor{lightgray}{\scriptsize{[ICLR22]}}\cite{hulora} & — & \checkmark & 89.46 & 85.5 & 78.26 & 83.78 & 56.20 & 72.25 \\
AdaptFormer \textcolor{lightgray}{\scriptsize{[NeurIPS22]}}\cite{chen2022adaptformer} & — & \checkmark & — & — & 80.56 & 84.88 & 58.83 & 72.32 \\
ARC$_{\text{att}}$ \textcolor{lightgray}{\scriptsize{[NeurIPS23]}}\cite{dong2023efficient} & — & \checkmark & 89.12 & 89.0 & 80.41 & 85.55 & 58.38 & 72.32 \\
\hline
VPT-S \textcolor{lightgray}{\scriptsize{[ECCV22]}}\cite{jia2022visual} & 0.16 & \checkmark & 84.62 & 85.5 & 76.81 & 79.66 & 46.98 & 64.85 \\
VPT-D \textcolor{lightgray}{\scriptsize{[ECCV22]}}\cite{jia2022visual} & 0.73 & \checkmark & 89.11 & 85.5 & 78.48 & 82.43 & 54.98 & 69.43 \\
E2VPT \textcolor{lightgray}{\scriptsize{[ICCV23]}}\cite{han20232} & 0.39 & \checkmark & 89.22 & 88.5 & 80.01 & 84.43 & 57.39 & 71.42 \\
EXPRES \textcolor{lightgray}{\scriptsize{[CVPR23]}}\cite{das2023learning} & — & \checkmark & — & — & 79.69 & 84.03 & 54.99 & 70.02 \\
DAM-VP \textcolor{lightgray}{\scriptsize{[CVPR23]}}\cite{huang2023diversity} & — & \checkmark & — & 88.5 & — & — & — & — \\
SA\textsuperscript{2}VP \textcolor{lightgray}{\scriptsize{[AAAI24]}}\cite{pei2024sa2vp} & 0.81 & \checkmark & \underline{90.08} & \underline{91.5} & \underline{80.97} & \textbf{85.73} & \underline{60.80} & \underline{75.83} \\
SPT \textcolor{lightgray}{\scriptsize{[ICML24]}}\cite{wangrevisiting} & — & \checkmark & 90.10 & — &  81.44 &  83.65 & 52.86 & 72.65 \\
VFPT \textcolor{lightgray}{\scriptsize{[NeurIPS24]}}\cite{zeng2024visual} & 0.66 & \checkmark & 89.24 & — & 81.35 & 84.93 & 60.19 & 75.49 \\
LoR-VP \textcolor{lightgray}{\scriptsize{[ICLR25]}}\cite{jin2025lor} & — & \checkmark & 91.22 & — & 80.25 & 85.12 & 58.71 & 74.69 \\
DA-VPT \textcolor{lightgray}{\scriptsize{[CVPR25]}}\cite{ren2025vpt} & — & \checkmark & 89.32 & — & 79.91 & 83.16 & 60.01 & 74.36 \\
\rowcolor{cvprblue!15}
\textbf{Ours} & 0.75 & \checkmark & \textbf{91.60} & \textbf{92.5} & \textbf{82.88} & \underline{85.61} & \textbf{62.55} & \textbf{77.01} \\
\Xhline{4\arrayrulewidth}
\end{tabular}
}
\label{tab:vit}
\vspace{-1.5em}
\end{table*}

\begin{table*}[!ht]
\centering
\caption{\textbf{Comparison of the fine-tuning methods under different pretraining paradigms.}
We report VTAB-1k~\cite{zhai2019large} accuracy using ViT-Base~\cite{dosovitskiyimage} as the frozen backbone, pretrained with MAE~\cite{he2022masked} and MoCo v3~\cite{chen2021empirical}, respectively.
}
\label{table:mae_moco}
\resizebox{0.9\textwidth}{!}{%
\begin{tabular}{c||r|rrr||r|rrr}
\Xhline{4\arrayrulewidth}
\rowcolor{gray!20}
Pretrained & \multicolumn{4}{c||}{MAE~\cite{he2022masked}} & \multicolumn{4}{c}{MoCo v3~\cite{chen2021empirical}} \\
\rowcolor{gray!20}
Method & Tuned(\%) & \textit{Natural} & \textit{Specialized} & \textit{Structured} & Tuned(\%) & \textit{Natural} & \textit{Specialized} & \textit{Structured} \\
\hline \hline
Full \textcolor{lightgray}{\scriptsize{[CVPR22]}}\cite{iofinova2022well} & 100.00 & \underline{59.31} & \textbf{79.68} & 53.82 & 100.00 & 71.95 & 84.72 & 51.98 \\
\hline
Linear \textcolor{lightgray}{\scriptsize{[CVPR22]}}\cite{iofinova2022well} & 0.04 & 18.87 & 53.72 & 23.70 & 0.04 & 67.46 & 81.08 & 30.33 \\
Partial-1 \textcolor{lightgray}{\scriptsize{[NeurIPS14]}}\cite{yosinski2014transferable} & 8.30 & 58.44 & 78.28 & 47.64 & 8.30 & 72.31 & 84.58 & 47.89 \\
\hline
Bias \textcolor{lightgray}{\scriptsize{[NeurIPS17]}}\cite{rebuffi2017learning} & 0.16 & 54.55 & 75.68 & \underline{47.70} & 0.16 & 72.89 & 81.14 & 53.43 \\
Adapter \textcolor{lightgray}{\scriptsize{[NeurIPS20]}}\cite{cai2020tinytl} & 0.87 & 54.90 & 75.19 & 38.98 & 1.12 & 74.19 & 82.66 & 47.69 \\
\hline
VPT-S \textcolor{lightgray}{\scriptsize{[ECCV22]}}\cite{jia2022visual} & 0.05 & 39.96 & 69.65 & 27.50 & 0.06 & 67.34 & 82.26 & 37.55 \\
VPT-D \textcolor{lightgray}{\scriptsize{[ECCV22]}}\cite{jia2022visual} & 0.31 & 36.02 & 60.61 & 26.57 & 0.22 & 70.27 & 83.04 & 42.38 \\
GPT \textcolor{lightgray}{\scriptsize{[ICML23]}}\cite{yoo2023improving} & 0.05 & 47.61 & 76.86 & 36.80 & 0.06 & 74.84 & 83.38 & 49.10 \\
VFPT \textcolor{lightgray}{\scriptsize{[NeurIPS24]}}\cite{zeng2024visual} & 0.38 & 53.59 & 77.75 & 36.15 & 0.22 & \underline{77.47} & \underline{85.76} & \underline{58.74} \\
LoR-VP \textcolor{lightgray}{\scriptsize{[ICLR25]}}\cite{jin2025lor} & 0.40 & 51.23 & 74.21 & 33.16 & 0.29 & 72.28 & 82.89 & 46.92 \\
DA-VPT \textcolor{lightgray}{\scriptsize{[CVPR25]}}\cite{ren2025vpt} & — & \textbf{62.14} & \underline{79.14} & \textbf{54.31} & — & 74.24 & 83.21 & 55.23 \\
\rowcolor{cvprblue!15}
\textbf{Ours} & 0.34 & 55.12 & 78.19 & 39.01 & 0.25 & \textbf{79.60} & \textbf{86.86} & \textbf{61.01} \\
\hline
\end{tabular}
}

\vspace{-2.0em}
\end{table*}

\subsection{Main Results}

\paragraph{ViT-Base.}
In Table~\ref{tab:vit}, we show that our method attains the best mean accuracy on ViT-Base across VTAB-1k, FGVC, and HTA. Specifically, relative to the fixed fusion VPT-Deep, gains are significant and consistent: VTAB-1k mean +7.58, FGVC +2.49, HTA +7.0, with category-wise improvements on \textit{Natural} +4.40, \textit{Specialized} +3.18, and \textit{Structured} +7.57, respectively. Compared with VFPT, we lift VTAB-1k mean by +1.52 and improve each VTAB category (\textit{Natural} +1.53, \textit{Specialized} $+0.68$, \textit{Structured} +2.36) at a comparable tunable budget (\(0.75\%\) $vs.$\ \(0.66\%\)), yielding a favorable accuracy–parameter trade-off (roughly \(+0.17\) points VTAB-mean per additional \(0.01\%\) tuned). Against SA$^{2}$VP, which is particularly strong on \textit{Specialized} (85.73\%), we obtain the top results on \textit{Natural} (82.88\%) and \textit{Structured} (62.55\%) and the best overall VTAB-1k mean (77.01\%, +1.18) with a similar parameter ratio. Taken together, these patterns indicate that \emph{how} prompts interact with tokens (\ie, the \emph{fusion rule} is a first-order factor for generalization): allowing each layer to choose between lightweight (\texttt{concat}/\texttt{add}) and semantic (\texttt{affine}/\texttt{cross-attn}) operators systematically reduces domain sensitivity (notably on \textit{Structured}) without resorting to heavier heads or hand-crafted inductive biases.

\paragraph{Different Pretrained Objectives.}
Table~\ref{table:mae_moco} shows that our fusion search is competitive under reconstruction-centric MAE and clearly strongest under contrastive MoCo v3. Under MAE, compared with VFPT (0.38\% tuned), we achieve higher accuracy in all VTAB-1k categories with a lower budget (0.34\%): \textit{Natural} 55.12 $vs.$\ 53.59 (+1.53), \textit{Specialized} 78.19 $vs.$\ 77.75 (+0.44), \textit{Structured} 39.01 $vs.$\ 36.15 (+2.86). Although the absolute category bests on MAE come from less parameter-efficient baselines (Partial-1 at 8.30\% on Natural/Specialized, Bias at 0.16\% on Structured), our results offer a balanced profile at PEFT scale, aligned with observation in~\cite{hanfacing}. Under MoCo v3, our method attains the top results across all categories while tuning only 0.25\%: \textit{Natural} 79.60 (+2.13 over VFPT at 0.22\%), \textit{Specialized} 86.86 (+1.10), and \textit{Structured} 61.01 (+2.27). The consistent gains over a strong frequency domain prompt at near identical budgets suggest that learning the fusion rule synergizes with contrastive features, likely by exploiting discriminative token relations via deeper affine/cross-attention choices. In short, across different pretrained objectives, we trade a modest tuned ratio for reliable cross-category improvements, narrowing MAE’s structured gap and delivering SOTA performance on MoCo v3.

\paragraph{Swin-Base.}
\begin{table}[t]
  \centering
  \scriptsize
  \setlength{\tabcolsep}{2.5pt}
  \renewcommand{\arraystretch}{0.86}
  \caption{
    \textbf{VTAB-1k results using Swin-Base.}
    Our method outperforms all baselines with fewer tunable parameters.
  }
  \label{tab:performance_vtab1k}
  \resizebox{0.6\textwidth}{!}{%
  \begin{tabular}{c||c|ccc}
    \Xhline{1pt}
    \rowcolor{gray!20}
    Swin-Base~\cite{liu2021swin} & Tuned & \textit{Nat.} & \textit{Spec.} & \textit{Struc.} \\
    \hline \hline
    Full~\cite{iofinova2022well} & 100.00 & 79.10 & 86.21 & 59.65 \\
    Linear~\cite{iofinova2022well} & 0.06 & 73.52 & 80.77 & 33.52 \\
    Bias~\cite{rebuffi2017learning} & 0.30 & 76.78 & 83.33 & 51.85 \\
    VPT-D~\cite{jia2022visual} & 0.25 & 76.78 & 83.33 & 51.85 \\
    E$^2$VPT~\cite{han20232} & 0.21 & 83.31 & 84.95 & 57.35 \\
    SA$^2$VP~\cite{pei2024sa2vp} & 0.29 & 80.81 & \underline{86.30} & \underline{60.03} \\
    VFPT~\cite{zeng2024visual} & 0.27 & \underline{84.53} & 86.15 & 58.21 \\
    LoR-VP~\cite{jin2025lor} & 0.29 & 83.51 & 85.22 & 57.61 \\
    \rowcolor{cvprblue!15}
    \textbf{Ours} & \textbf{0.26} & \textbf{85.52} & \textbf{86.49} & \textbf{62.36} \\
    \Xhline{1pt}
  \end{tabular}
  }
\end{table}

Table~\ref{tab:performance_vtab1k} shows that our fusion search remains effective on Swin-Base~\cite{liu2021swin}. Specifically, we achieve the best results in all VTAB-1k categories, surpassing VFPT at 0.27\% and SA$^{2}$VP at 0.29\%. Gains over VPT-Deep are substantial despite a nearly identical budget (0.26\% $vs.$ 0.25\%): \textit{Natural} +8.74\%, \textit{Specialized} +3.16\%, \textit{Structured} +10.51\%. The largest margins are seen on the \textit{Structured}, indicating that the layer-wise operator selection effectively adapts to Swin's stage-wise shifts and transitions from local to global reasoning. In the early blocks, lightweight fusion is favored to maintain locality within windows, while the deeper blocks choose more complex operators to integrate evidence across different windows. Additionally, across stage transitions, affine calibration helps stabilize the scales of the tokens. This schedule yields stronger geometry and counting-heavy performance without resorting to heavier per-layer heads, explaining why we simultaneously reduce domain sensitivity and maintain a competitive tunable ratio.

\subsection{Ablation Studies}
\label{sec:ablation}

\begin{table}[t]
  \centering
  \scriptsize
  \setlength{\tabcolsep}{2.5pt}
  \renewcommand{\arraystretch}{0.86}
  \caption{
    \textbf{Prompt fusion strategy ablation.}
    Search-based fusion consistently outperforms any single fixed operator.
  }
  \label{tab:fusion_ablation}
  \resizebox{0.6\textwidth}{!}{%
  \begin{tabular}{l||c|ccc|c}
    \Xhline{1pt}
    \rowcolor{gray!20}
    Fusion & Search & \textit{Nat.} & \textit{Spec.} & \textit{Struc.} & Gain \\
    \hline \hline
    \texttt{Concat}      & \textcolor{red}{\ding{55}} & 78.42 & 82.10 & 56.33 & -1.73 \\
    \texttt{Add}         & \textcolor{red}{\ding{55}} & 77.63 & 81.47 & 55.98 & -2.32 \\
    \texttt{Affine}      & \textcolor{red}{\ding{55}} & 78.86 & 82.62 & 57.40 & -1.05 \\
    \texttt{Cross-Attn}  & \textcolor{red}{\ding{55}} & \underline{79.91} & \underline{83.44} & \underline{58.67} & 0.00 \\
    \rowcolor{cvprblue!15}
    \textbf{Ours} & \textcolor{naturegreen}{\ding{51}} & \textbf{82.88} & \textbf{85.61} & \textbf{62.55} & \textbf{+3.00} \\
    \Xhline{1pt}
  \end{tabular}
  }
\end{table}

Table \ref{tab:fusion_ablation} varies only the fusion rule while holding the prompt length and backbone fixed. Among single operators, cross-attention is the strongest. Our searchable fusion improves to 82.88 / 85.61 / 62.55 on Natural / Specialized / Structured, \ie, +2.97, +2.17, and +3.88 over the best fixed baseline, for a +3.00 mean gain across VTAB-1k groups. Concat, Add, and Affine trail cross-attention by 1.73, 2.32, and 1.05 mean points, respectively, indicating that no single operator is uniformly optimal. The largest margin on Structured suggests that layer-wise operator choice is particularly helpful for token calibration and geometric reasoning. Overall, the results validate our hypothesis: learning the fusion rule via per layer selection delivers consistent, task robust gains beyond improving prompt content alone.

\subsection{Layer-wise Fusion Patterns}
\label{sec:analysis}

\begin{figure}[t]
  \centering
  \includegraphics[width=0.72\linewidth]{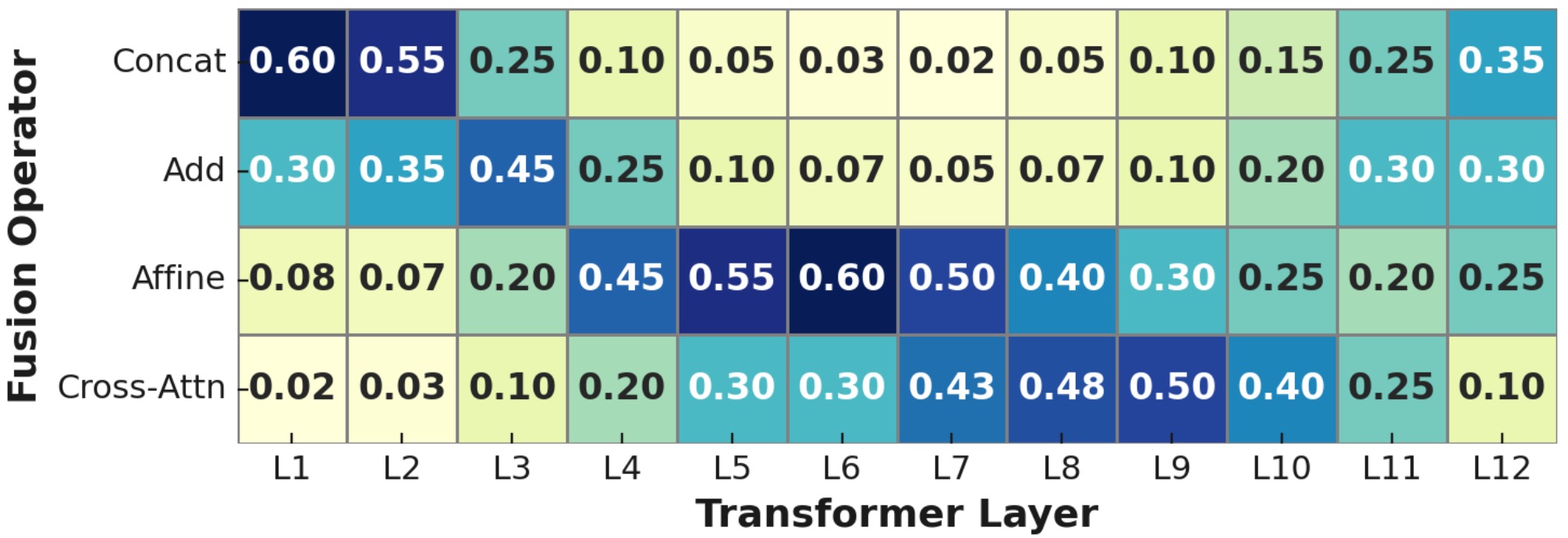}
  \caption{\textbf{Fusion operator selection by layer.}
  Early layers favor \texttt{concat} and \texttt{add}, while deeper layers shift toward \texttt{affine} and \texttt{cross-attention}.}
  \label{fig:fusion_softmax_heatmap}
\end{figure}

\paragraph{Depth-wise Patterns.}
Figure~\ref{fig:fusion_softmax_heatmap} shows a clear depth profile. Early blocks concentrate probability mass on \texttt{concat}/\texttt{add}. These operators minimally perturb token statistics and act as content-preserving routes, allowing prompts to tag the stream without overwriting low-level evidence. As depth increases, the search shifts toward \texttt{affine}/\texttt{cross-attention}. The former calibrates feature scales and offsets (helpful when domain statistics drift), while the latter enables content-adaptive routing between prompts and tokens. Aggregated over VTAB-1k, layers 1--4 assign $0.69$ probability mass to lightweight operators, layers 5--8 have the highest entropy, and layers 9--12 assign $0.73$ mass to \texttt{affine}/\texttt{cross-attention}. The final operators agree on 9--10 of 12 layers across three seeds, with most variation concentrated in the middle blocks. Consistent with the ablation, the layers that finally prefer \texttt{cross-attention} are also those whose replacement with a single fixed rule produces the most significant accuracy drop, suggesting that \emph{where} we switch operators is as important as \emph{which} operator we pick.

\begin{figure}[t]
  \centering
  \includegraphics[width=0.8\columnwidth]{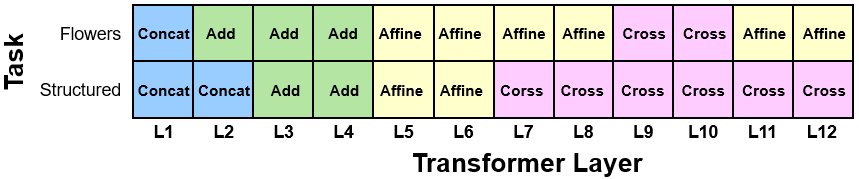}
  \caption{\textbf{Task-specific fusion preferences.}
  The discovered operator distributions differ between Flowers and Structured tasks: early layers consistently favor lightweight fusion (\texttt{Concat}/\texttt{Add}), while deeper layers shift toward semantic operators (\texttt{Affine}/\texttt{Cross}), revealing task-specific adaptation patterns.}
  \label{fig:task_fusion_comparison}
\end{figure}

\paragraph{Task-dependent Preferences.}
Figure~\ref{fig:task_fusion_comparison} compares architectures discovered on FGVC versus Structured regimes. FGVC consistently pushes \texttt{cross-attention} deeper, which is coherent with the need to bind subtle, instance-specific attributes to category cues. Structured tasks select \texttt{cross-attention}/\texttt{affine} more often than Natural tasks, reflecting a stronger emphasis on geometric calibration and token rescaling. Any static recipe does not capture these task-aware shifts and helps explain the empirical pattern in Table~\ref{tab:fusion_ablation}: the most significant margin over the best fixed rule appears on the Structured group, where per-layer calibration is particularly valuable.

\begin{figure}[t]
  \centering
  \includegraphics[width=0.9\linewidth]{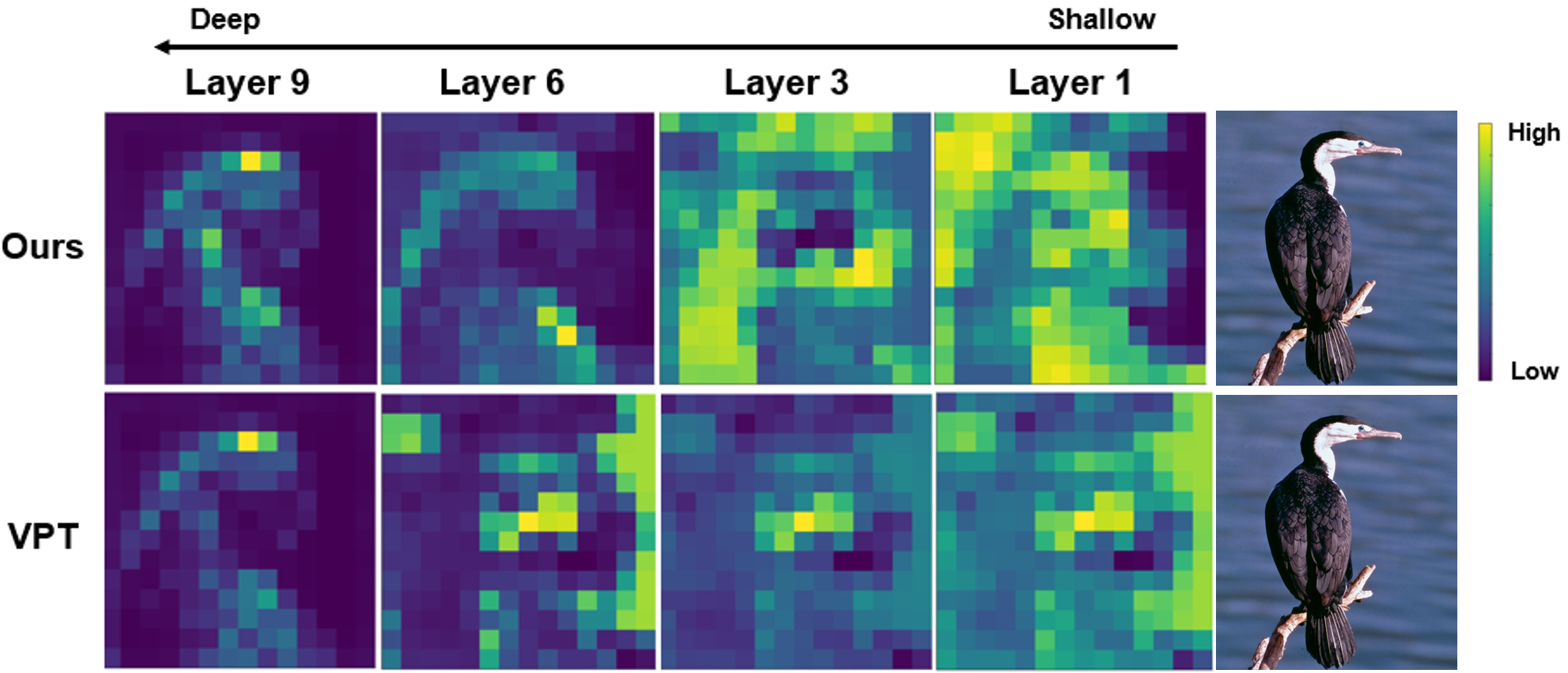}
  \caption{\textbf{Layer-wise attention visualization.}
  We compare CLS$\rightarrow$patch attention (mean over heads) at four representative layers.
  Each column uses the same color scale for a fair comparison. Our model shows more concentrated attention over object regions in deeper layers compared to VPT \cite{jia2022visual}.}
  \label{fig:attn_grid}
\end{figure}

\paragraph{Layer-wise Attention Analysis.}
We select four representative layers and visualize the CLS$\rightarrow$patch attention for VPT \cite{jia2022visual} and our model (Figure~\ref{fig:attn_grid}). At layer 1, both models distribute attention broadly. By layer 3, our model begins to concentrate attention on the head region, while VPT still shows several scattered peaks. At layer 6, our model assigns less attention to peripheral patches near the right boundary and more to the eye/beak area, whereas VPT continues to spread attention across multiple regions. By layer 9, our model creates a compact hotspot on the eye, while VPT remains more diffuse. These observations should be read as attention-concentration evidence rather than proof of literal background removal: background can be task-relevant, and cross-attention may also act by aggregating prompt-token evidence. Across images, attention entropy decreases with depth and object-region energy increases for our model \cite{heo2021rethinking}.

\subsection{Analysis of Efficiency and Inference Latency}
\label{sec:efficiency_latency}

\begin{figure}[t]
  \centering
  \includegraphics[width=0.9\linewidth]{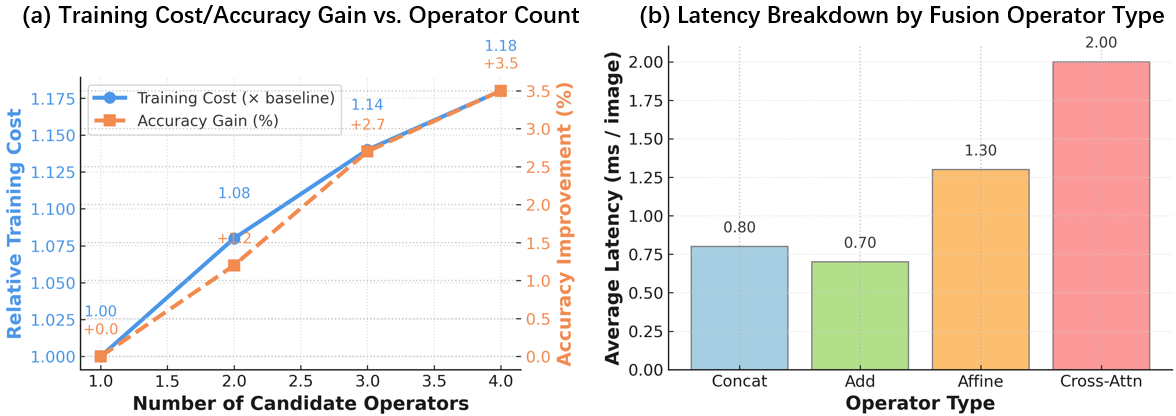}
  \caption{\textbf{Efficiency analysis.}
  (a)~Training cost scales sublinearly with more candidate operators,
  while accuracy steadily improves.
  (b)~Latency breakdown shows lightweight \texttt{add}/\texttt{concat}
  and moderate overhead from \texttt{affine}/\texttt{cross-attention}. We use ViT-B/16 as backbone and a batch size of 64 on an NVIDIA A100 GPU.}
  \label{fig:efficiency_combined}
\end{figure}

To evaluate the deployment cost of our method, we measure both the training overhead during the search stage and the inference latency after discretization on VTAB-1k.
During search, each transformer block evaluates a soft mixture of candidate fusion operators, which adds a few extra forward passes proportional to the number of candidates.
As shown in Figure~\ref{fig:efficiency_combined}(a), the training cost increases slowly with more operators. With four candidates, the search cost is $1.38\times$ that of VPT-Deep~\cite{jia2022visual} (10.8 vs.\ 7.8 GPU hours), while the VTAB-1k mean improves by 3.5 points. This extra cost mainly comes from deeper layers that evaluate \texttt{cross-attention} during search. After discretization, inference uses only one operator per layer. Our model runs at 15.9 ms per image compared with 14.8 ms for VPT-Deep, a 7.4\% increase. Under matched controls, latency-matched and parameter-matched variants still reach 75.83\% and 76.42\% VTAB-1k mean, respectively, while VPT-Deep reaches 69.43\%; full numbers are in Appendix~\ref{app:space}. As shown in Figure~\ref{fig:efficiency_combined}(b), \texttt{add} and \texttt{concat} are very fast, while \texttt{affine} and \texttt{cross-attention} take longer, matching their computational complexity. The efficiency comes from late-stage discretization, which prunes inactive fusion paths, and from operator heterogeneity, which lets shallow layers use cheap operations while deeper layers use more expressive ones only when needed.

\section{Conclusion}
\label{sec:conclusion}

In this paper, we challenge the common practice of using a fixed operator to combine prompts and image tokens in visual prompt tuning. We propose a novel approach that treats fusion as a learnable component, jointly optimizing the prompt parameters and their mode of fusion through differentiable architecture search. This exploration encompasses four fusion methods: concatenation, addition, affine transformation, and cross-attention. This allows us to discover diverse and layer-specific fusion techniques that adapt to the representational needs of the frozen backbone. By introducing fusion operator search into the prompt tuning paradigm, our work provides a fresh perspective on understanding and enhancing prompt utilization. This work bridges the gap between neural architecture search and lightweight fine-tuning, presenting an efficient approach to designing prompting solutions for large-scale vision models.

\makeatletter
\let\@vspace\old@vspace
\let\@vspacer\old@vspacer
\makeatother

\section*{Acknowledgements}
This manuscript was co-authored by Oak Ridge National Laboratory (ORNL), operated by UT-Battelle, LLC under Contract No. DE-AC05-00OR22725 with the U.S. Department of Energy.  
Any subjective views or opinions expressed in this paper do not necessarily represent those of the U.S. Department of Energy or the United States Government. This research was supported by the National Science Foundation under Grant No. 2450068.

\bibliographystyle{unsrtnat}
\bibliography{main}

\clearpage

\renewcommand{\thefigure}{S\arabic{figure}}
\renewcommand{\theHfigure}{S\arabic{figure}}
\setcounter{figure}{0}
\renewcommand{\thetable}{S\arabic{table}}
\renewcommand{\theHtable}{S\arabic{table}}
\setcounter{table}{0}
\renewcommand{\theHsection}{\Alph{section}}
\renewcommand{\theHsubsection}{\Alph{section}.\arabic{subsection}}
\renewcommand{\theHsubsubsection}{\Alph{section}.\arabic{subsection}.\arabic{subsubsection}}

\clearpage
\newpage

\appendix

\section*{Overview of Appendix}

This supplementary material presents additional theoretical derivations, detailed experimental results, and comprehensive efficiency analyses to support the main paper. The document is organized as follows:

\begin{itemize}

    \item \textbf{Section 5} related work.

    \item \textbf{Section 6} provides the mathematical derivations for the differentiable search, including the implicit gradient approximation and the properties of the fusion operators. It also details the protocol used for Mutual Information estimation.

    \item \textbf{Section 7} reports the full per-task performance metrics across the FGVC, HTA, and VTAB-1k benchmarks, supplementing the aggregated results in the main paper.

    \item \textbf{Section 8} describes the specific architectural implementations of the fusion operators (Concat, Add, Affine, and Cross-Attention), validates efficiency-oriented design choices, and analyzes prompt-summary conditioning.

    \item \textbf{Section 9} elaborates on the search stability mechanisms, including the temperature annealing schedule, regularization coefficients, and gradient clipping strategies.

    \item \textbf{Section 10} analyzes search-space variants, matched-budget controls, warm-start compatibility, and preliminary dense-prediction extensions.

    \item \textbf{Section 11} investigates the layer-wise patterns, specifically the correlations between operator selection and information-theoretic metrics across different depths.

    \item \textbf{Section 12} provides an extended analysis of system efficiency, covering latency breakdowns, memory consumption, and scaling properties on various hardware configurations.

    \item \textbf{Section 13} discusses the limitations of the current approach and outlines potential directions for future research.
\end{itemize}

\section{Related Work}
\label{sec:related}

\subsection{Parameter-Efficient Fine-Tuning in Vision}
Parameter–Efficient Fine–Tuning (PEFT) adapts large pretrained models by updating a small set of task–specific parameters while keeping the backbone frozen. Existing methods are commonly grouped into three categories: (i) \emph{extra modules}, which insert lightweight blocks into the network, including Adapters and their variants~\cite{houlsby2019parameter,rebuffi2017learning,pfeiffer2020adapterhub}, Side–Tuning~\cite{zhang2020side}, and compact/bottleneck designs such as Compacter, TinyTL, and AdaptFormer~\cite{karimi2021compacter,cai2020tinytl,chen2022adaptformer}; (ii) \emph{reparameterization}, which modifies weight updates with low‑rank adapters, typified by LoRA~\cite{hulora}; and (iii) \emph{prompt–based} tuning, which conditions a frozen backbone through learned soft vectors or sequences~\cite{li2021prefix,lester2021power,shin2020autoprompt,wallace2019universal,dettmers2023qlora,karimi2021compacter,zaken2022bitfit, yan2023prompt}. In vision, these strategies deliver strong accuracy–efficiency trade‑offs on top of ViTs. A representative line is \emph{Visual Prompt Tuning} (VPT)~\cite{jia2022visual}, which inject learnable tokens to the patch sequence and optimizes only the prompt tokens. Subsequent research explore where and how to place prompts across depth~\cite{jia2022visual,yoo2023improving,ren2025vpt,zeng2024visual,xiao2025visual,xiao2025visualv,wangrevisiting,yoo2023improving,shang2025pro,ren2025vpt}, design spatially structured or diversity‑aware prompts~\cite{pei2024sa2vp,huang2023diversity}, introduce frequency‑domain variants~\cite{zeng2024visual}, and improve interaction/efficiency with decoupled or residual formulations and cross‑attention couplings~\cite{han20232,das2023learning,huang2025cvpt}. Despite the breadth of designs, most works \emph{fix} the fusion scheme between prompts and tokens throughout the network, typically used scheme is either concatenation or addition~\cite{jia2022visual}. Our work focuses on this missing piece: we revisit fusion as a first‑class, searchable component and allow each layer to select among a small set of basic operators while keeping the ViT backbone frozen and the interface unchanged.

\subsection{Neural Architecture Search for ViTs}
Differentiable Neural Architecture Search (NAS) relaxes discrete choices into a continuous space and optimizes them with gradient‑based bilevel objectives~\cite{liudarts,elsken2019neural}. In vision, NAS has been used to design Transformer blocks, widths, and heads~\cite{chen2021autoformer,su2022vitas}, to stabilize supernets~\cite{gong2021nasvit}, and to improve efficiency via dynamic token routing or early exiting~\cite{rao2021dynamicvit,liangevit}. In~\cite{li2025advancing}, NAS is used to search semantic attributes that are implemented in textual information,~\cite{huang2025differentiable} automates the continuous prompt learning process. It automatically determines the continuous prompts and associated hyperparameters including the context length of the continuous prompt inserted for
each neural network layer and prompt depth.
We place prompt tuning inside this NAS framework: instead of committing to a single fusion mechanism, we search over a compact set of basic operators with a differentiable relaxation, yielding layer‑wise operator choices that better match task and depth while keeping the frozen ViT interface unchanged.

\section{Additional Derivations and Proofs}\label{app:proofs}

This section provides derivations of the implicit gradient used in the bilevel search, details of the Hessian–vector product computation, gradients of regularizers and temperature, properties of the fusion operators (convex reduction and identity preservation), complexity bounds, and the mutual-information estimation protocol.

Regarding the nomenclature of our search space, we follow the established conventions in previous work~\cite{jia2022visual, hanfacing}, where the fundamental fusion mechanisms between prompts and image tokens are categorized under standard terms such as `concat' and `add'. While we acknowledge that these operators in our framework incorporate specific enhancements (e.g., the token-preserving mixer for \texttt{concat} and the mean-bias injection for \texttt{add}), we maintain these names to ensure consistency with baseline visual prompt tuning methods.

\subsection{Preliminaries and Notation}
We follow the notation in Sec.~\ref{sec:method}.
For layer $l$, $\alpha^{(l)}\in\mathbb{R}^{|\mathcal S|}$ are architecture logits,
$\pi^{(l)}=\mathrm{softmax}(\alpha^{(l)}/\tau)$ are mixing weights,
and the soft fusion is
\begin{equation}\label{eq:supp-soft}
\Delta^{(l)}_{\text{soft}}(p^{(l)},\tilde x^{(l-1)})
= \sum_{i\in\mathcal S}\pi^{(l)}_i\,\Delta_i\!\big(p^{(l)},\tilde x^{(l-1)}\big).
\end{equation}
The inner/outer losses are $\mathcal L_{\mathrm{train}}(\phi,\alpha)$ and
$\mathcal L_{\mathrm{val}}(\phi,\alpha)$, and we use the regularizers in Eq.~\ref{eq:reg}.

\subsection{Bilevel Optimization: Implicit Gradient and Unrolled Approximation}
\label{app:implicit}

\paragraph{Implicit differentiation.}
Let $\phi^\ast(\alpha)=\arg\min_\phi \mathcal L_{\mathrm{train}}(\phi,\alpha)$ and
$F(\phi,\alpha)\!=\!\nabla_\phi \mathcal L_{\mathrm{train}}(\phi,\alpha)$.
At an inner optimum $F(\phi^\ast,\alpha)=0$. Differentiating w.r.t.\ $\alpha$ and solving for
$\mathrm{d}\phi^\ast/\mathrm{d}\alpha$ gives
\begin{equation}
\nabla_{\alpha}\phi^\ast(\alpha)
= - \big[\nabla^2_{\phi\phi}\mathcal L_{\mathrm{train}}(\phi^\ast,\alpha)\big]^{-1}
  \nabla^2_{\alpha\phi}\mathcal L_{\mathrm{train}}(\phi^\ast,\alpha).
\end{equation}
Hence the implicit gradient of the outer objective is
\begin{equation}
\label{eq:supp-implicit}
\begin{aligned}
\nabla_\alpha \mathcal L_{\mathrm{val}}\big(\phi^\ast(\alpha),\alpha\big)
&= \underbrace{\nabla_\alpha \mathcal L_{\mathrm{val}}}_{\text{direct}}
\\[-2pt]
&\quad -\;
\underbrace{\nabla^2_{\alpha\phi}\mathcal L_{\mathrm{train}}\,
\Big[\nabla^2_{\phi\phi}\mathcal L_{\mathrm{train}}\Big]^{-1}\,
\nabla_\phi \mathcal L_{\mathrm{val}}}_{\text{response term}}\,.
\end{aligned}
\end{equation}

\paragraph{One-step unrolled approximation (DARTS).}
Following~\cite{liudarts}, we take a single inner-step
\[
\phi'=\phi-\eta_\phi \nabla_\phi \mathcal L_{\mathrm{train}}(\phi,\alpha)
\]
and use
\begin{equation}
\label{eq:supp-unroll}
\begin{aligned}
\nabla_{\alpha}\mathcal{L}_{\mathrm{val}}(\phi',\alpha)
&\approx \nabla_{\alpha}\mathcal L_{\mathrm{val}}(\phi',\alpha)\\
&\quad - \eta_\phi\, \Big[\nabla^2_{\alpha\phi}\mathcal L_{\mathrm{train}}(\phi,\alpha)\Big]\,
           \nabla_\phi \mathcal L_{\mathrm{val}}(\phi',\alpha).
\end{aligned}
\end{equation}

which is Eq.~\ref{eq:unroll} in the main paper.

\paragraph{Hessian--vector product (HVP).}
Let $H=\nabla^2_{\phi\phi}\mathcal L_{\mathrm{train}}(\phi,\alpha)$.
For any vector $v$ in the parameter space of $\phi$,
\begin{equation}
\label{eq:hvp-identity}
Hv \;=\; \nabla_\phi \Big(\,\nabla_\phi \mathcal L_{\mathrm{train}}(\phi,\alpha)^\top v\,\Big).
\end{equation}
Eq.~\eqref{eq:hvp-identity} yields an $O(\text{backprop})$-time product without forming $H$.
In practice, HVP is computed by two reverse-mode passes:
\begin{enumerate}[leftmargin=1.2em,itemsep=1pt,topsep=2pt]
\item First backward (with graph): $g_\phi=\nabla_\phi \mathcal L_{\mathrm{train}}(\phi,\alpha)$.
\item Take a scalar dot: $s=\langle g_\phi, v\rangle$.
\item Second backward: $Hv=\nabla_\phi s$.
\end{enumerate}
This avoids explicit Hessian materialization; memory overhead comes from retaining the graph for the second pass.

\paragraph{Mixed Hessian--vector product.}
The response term in Eq.~(S\ref{eq:supp-implicit}) contains $\nabla^2_{\alpha\phi}\mathcal L_{\mathrm{train}}$.
Given $u$ in the $\phi$-space,
\begin{equation}
\label{eq:mixed-hvp}
\big(\nabla^2_{\alpha\phi}\mathcal L_{\mathrm{train}}\big)\,u
\;=\;
\nabla_\alpha\,\Big(\,\nabla_\phi \mathcal L_{\mathrm{train}}(\phi,\alpha)^\top u\,\Big).
\end{equation}
It is the first form the inner product $\langle\nabla_\phi \mathcal L_{\mathrm{train}},u\rangle$, then backpropagate with respect to $\alpha$ to obtain the mixed HVP without constructing any Hessian explicitly.

\paragraph{Linear solve for implicit schemes.}
When an inverse is needed (e.g., solving $Hz=g$ in implicit differentiation), we use a few iterations of damped conjugate gradients (CG) with the HVP oracle $v\mapsto Hv$:
\begin{equation}
\label{eq:cg-damped}
(H+\lambda I)\,z=g,\qquad v\mapsto Hv+\lambda v,
\end{equation}
with small damping $\lambda>0$, tolerance $10^{-3}$, and at most 5--10 iterations.
A diagonal preconditioner $M\approx\mathrm{diag}(H)$ (estimated from Fisher/gradient variances) further stabilizes the solve.
DARTS in our work uses the one-step unrolled surrogate (Eq.~(S\ref{eq:supp-unroll})), so the full implicit solve is not required in the main training loop; we include Eq.~\eqref{eq:cg-damped} for completeness.

\paragraph{Implementation notes.}
(i) The first backward must build a graph (e.g., \texttt{create\_graph=True}) so the second backward can traverse it;
(ii) detach $v$ to avoid unwanted gradients;
(iii) clear/reuse graphs carefully to control memory;
(iv) for $\nabla^2_{\alpha\phi}$–vector products use Eq.~\eqref{eq:mixed-hvp} by differentiating the dot scalar with respect to $\alpha$.

\subsection{Gradients of the Regularizers and Temperature}
\label{app:reg-grad}
For $\pi=\mathrm{softmax}(\alpha/\tau)$,
\begin{equation}
\frac{\partial \pi_i}{\partial \alpha_j}=\frac{1}{\tau}\,\pi_i\,(\mathbf 1_{i=j}-\pi_j).
\end{equation}
Let $H(\pi)=-\sum_i \pi_i \log \pi_i$. Then
\begin{equation}
\begin{aligned}
\frac{\partial H}{\partial \alpha_j}
&= -\sum_i (\log\pi_i + 1)\,\frac{\partial \pi_i}{\partial \alpha_j}\\
&= -\frac{1}{\tau}\sum_i \pi_i\,(\delta_{ij}-\pi_j)\,(\log\pi_i + 1).
\end{aligned}
\end{equation}
Here $\delta_{ij}$ is the Kronecker delta (1 if i=j, 0 otherwise).

The cost prior gradient is
\begin{equation}
\frac{\partial}{\partial \alpha_j}\sum_i c_i\pi_i
= \sum_i c_i\,\frac{\partial \pi_i}{\partial \alpha_j}
= \frac{1}{\tau}\,\pi_j\,(c_j-\sum_i c_i\pi_i).
\end{equation}
As $\tau\!\downarrow\!0$, $\pi$ concentrates on $\arg\max_i \alpha_i$; rigorously, for unique maximizer $i^\star$, $\lim_{\tau\to 0}\pi_{i^\star}=1$ and $\lim_{\tau\to 0}\pi_{i}=0$ for $i\neq i^\star$.

\subsection{Properties of Fusion Operators}
\label{app:ops-prop}

\paragraph{(a) Token-preserving mixer is a convex reducer.}
Let $R^{(l)}=\mathrm{softmax}_{\mathrm{col}}(U^{(l)})$, so each column $R^{(l)}_{:,j}$ is nonnegative and sums to $1$. Then for $\widetilde x=[p^{(l)};\tilde x^{(l-1)}]$,
\begin{equation}
\big(R^{(l)}\big)^\top \widetilde x
=\begin{bmatrix}
\sum_{t=1}^{m+k} R^{(l)}_{t,1}\,\widetilde x_t\\
\vdots\\
\sum_{t=1}^{m+k} R^{(l)}_{t,k}\,\widetilde x_t
\end{bmatrix},
\end{equation}
so every output token is a convex combination of $(m{+}k)$ inputs (no projections $W_Q/W_K/W_V$ are used).

\paragraph{(b) Identity preservation.}
If prompts are uninformative, the operators can reduce to identity:
\begin{equation}
\begin{aligned}
\texttt{add}:~&s=0 \;\Rightarrow\; \Delta_{\texttt{add}}=\tilde x^{(l-1)},\\
\texttt{affine}:~&\gamma=\mathbf 1,\ \beta=\mathbf 0 \;\Rightarrow\; \Delta_{\texttt{affine}}=\tilde x^{(l-1)},\\
\texttt{concat}:~&R^{(l)}=\begin{bmatrix}0_{m\times k}\\ I_k\end{bmatrix}\ \Rightarrow\ \Delta_{\texttt{concat}}=\tilde x^{(l-1)},\\
\texttt{cross}:~&A=0 \ \text{(or }V_p=0)\ \Rightarrow\ \Delta_{\texttt{cross}}=\tilde x^{(l-1)}.
\end{aligned}
\end{equation}

\paragraph{(c) Why inject after LN (Pre-LN blocks).}
For any vector $z$, $\mathrm{LN}(z)=\frac{z-\mu(z)}{\sigma(z)}$ (elementwise), thus
$\mathrm{LN}(z+b)=\frac{z-\mu(z)}{\sigma(z)}$ for any constant bias $b$; additive/affine modulations applied \emph{before} LN are removed by normalization.
Injecting $\Delta(\mathrm{LN}(\cdot))$ \emph{before the frozen MSA/MLP and without another LN afterwards} preserves the modulation, as used in Sec.~\ref{sec:method}.

\paragraph{(d) Discretization limit.}
With unique maximizer $i^\star=\arg\max_i \alpha^{(l)}_i$, the soft operator in Eq.~\ref{eq:supp-soft} converges to the hard choice:
\begin{equation}
\lim_{\tau\to 0}\Delta^{(l)}_{\text{soft}}
= \Delta_{i^\star}\big(p^{(l)},\tilde x^{(l-1)}\big).
\end{equation}

\paragraph{Complexity Bounds and Parameter Counts}
Let $k$ be the token count, $m$ the prompt length, $d$ the hidden size, $H$ the number of heads, and $d_h{=}d/H$. We estimate
\[
\mathrm{cost}(\cdot)=
\begin{cases}
O(kd), & \texttt{add}\\
O(kd)+O(d^2), & \texttt{affine}\\
O((m{+}k)kd), & \texttt{concat}\\
O(Hkm\,d_h)+O((k{+}m)d^2), & \texttt{cross}
\end{cases}
\]
where the $O(Hkm\,d_h)$ term counts $QK^\top V$ and the $O((k{+}m)d^2)$ term counts the $Q/K/V$ projections (shared across heads). In typical ViT settings (\(d\) large and \(m{\ll}k\)), \texttt{cross} is often \emph{heavier} than \texttt{concat} because the projection term \(O((k{+}m)d^2)\) dominates, while \texttt{add}/\texttt{affine} remain the cheapest. We define a nonnegative cost vector $c=[c_{\texttt{add}},c_{\texttt{affine}},c_{\texttt{concat}},c_{\texttt{cross}}]$, normalized from microbenchmarks, and include $\sum_{l,i} c_i\,\pi^{(l)}_i$ as a soft regularizer in the outer objective (Eq.~\ref{eq:reg}). Heavier operators are not always better, but they can be essential at specific depths; we therefore keep them in $\mathcal{S}$ and rely on data-driven selection with late discretization to activate them only when they bring consistent gains.

\subsection{Mutual Information Estimation Details}
\label{app:mi}

We estimate each layer mutual information with three standard variational estimators and report consistent trends across them. For layer $l$, the representation is $T^{(l)}\!\in\!\mathbb{R}^{d_T}$ (CLS token after the block’s pre-LN unless otherwise noted). We consider two pairs: $(T^{(l)},X)$ and $(T^{(l)},Y)$, where $X$ is the input image and $Y$ is the label.

\paragraph{Estimators.}
Let $(U,V)$ denote a generic pair. We implement:

\noindent\textbf{(a) MINE (DV bound).}
A neural critic $f_\theta(U,V)$ optimizes the Donsker--Varadhan lower bound
\begin{equation}
\begin{aligned}
\widehat I_{\mathrm{MINE}}
&= \mathbb{E}_{p(u,v)}\!\big[f_\theta(u,v)\big]
 - \log \mathbb{E}_{p(u)p(v)}\!\big[\exp f_\theta(u,v)\big].
\end{aligned}
\end{equation}
We use an exponential moving average for the second term to stabilize training.

\noindent\textbf{(b) InfoNCE.}
Using $K$ negatives per positive in a batch,
\begin{equation}
\begin{aligned}
\widehat I_{\mathrm{NCE}}
&= \mathbb{E}\Bigg[
\log \frac{\exp s(u,v)}{\sum_{j=1}^{K}\exp s(u,v_j)}
\Bigg],
\end{aligned}
\end{equation}
with a score $s(u,v)$.

\noindent\textbf{(c) CLUB (upper bound).}
A variational conditional model $q_\psi(v|u)$ yields
\begin{equation}
\begin{aligned}
\widehat I_{\mathrm{CLUB}}
&= \mathbb{E}_{p(u,v)}\!\big[\log q_\psi(v|u)\big]
 - \mathbb{E}_{p(u)p(v)}\!\big[\log q_\psi(v|u)\big],
\end{aligned}
\end{equation}
which is an upper bound when $q_\psi$ is optimized to fit $p(v|u)$.

\paragraph{Representations and pairing.}
We standardize $T^{(l)}$ by per-dimension mean/variance computed on the training split (frozen for evaluation). For $X$, we use resized RGB and pass it through a fixed random Fourier projection to reduce dimension:
\begin{equation}
\begin{aligned}
\phi(X)=
\sqrt{\tfrac{2}{D}}\,
\big[\cos(\Omega X + b),\ \sin(\Omega X + b)\big]\in\mathbb{R}^{D},
\end{aligned}
\end{equation}
with $\Omega\!\sim\!\mathcal N(0,\sigma^2 I)$, $b\!\sim\!\mathrm{Unif}[0,2\pi)$, $D{=}256$ (seed fixed across runs). This avoids training an extra encoder for $X$ and reduces variance. For $(T^{(l)},Y)$, $Y$ is the one-hot label (or smoothed label) and we feed $(T^{(l)},Y)$ directly into the estimator heads.

\paragraph{Training and evaluation protocol.}
Estimators are trained \emph{after} the backbone is frozen and the fusion is discretized; they do not update the backbone or prompts. We subsample $N_{\mathrm{MI}}$ images per dataset and compute MI independently per layer.
\begin{itemize}[leftmargin=1.2em, itemsep=2pt, topsep=2pt]
\item \textbf{Batches.} Balanced class sampling (when labels are available) and fixed negative pool size $K$ for InfoNCE.
\item \textbf{Critic heads.} $f_\theta$ and $q_\psi$ are two-layer MLPs with hidden size 512, SiLU activation, LayerNorm, dropout 0.1.
\item \textbf{Optimization.} AdamW, lr $1\text{e-}3$, weight decay $1\text{e-}4$, 5{,}000 steps per layer (early stop on a held-out MI-validation split of the same $N_{\mathrm{MI}}$ pool).
\item \textbf{Stabilization.} Gradient clip at 1.0; MINE uses moving-average coefficient 0.99 for the $\log \mathbb E[\exp(\cdot)]$ term; InfoNCE temperature $\tau_{\text{nce}}{=}0.07$.
\end{itemize}
We report $\widehat I(T^{(l)};X)$ and $\widehat I(T^{(l)};Y)$ as the average over three seeds with $95\%$ confidence intervals (t-interval). We also report the IB surrogate $\widehat{\mathcal L}^{(l)}_{\mathrm{IB}}=\widehat I(T^{(l)};X)-\beta\,\widehat I(T^{(l)};Y)$ with $\beta{=}1$ unless stated.

\paragraph{Prompt-conditioned compression.}
Operators in $\mathcal{S}$ induce different IB trade-offs:
\texttt{add}/\texttt{affine} act as channel-wise gates that suppress nuisance dimensions (tending to lower $\widehat{I}(T;X)$);
\texttt{concat} mixes prefix prompt cues into a token-preserving summary;
\texttt{cross-attention} performs content-adaptive retrieval from a prompt memory and emphasizes label-relevant patterns.
Learning the operator choice per layer lets shallow blocks prefer lightweight modulation and deeper blocks perform semantic integration, which matches the depth trends in Sec.~\ref{sec:analysis}.

\begin{algorithm}[t]
\caption{Layer-wise Search and Discretization for Prompt--Token Fusion}
\label{alg:search_appendix}
\begin{algorithmic}[1]
\Require Frozen ViT $f_\psi$; operator set $\mathcal S$; cost prior $c=\{c_i\}$; train/val splits
\Require Init prompts $\{p^{(l)}\}$ and operator-internal weights in $\phi$; architecture logits $\alpha=\{\alpha^{(l)}\}$
\Require Hyperparams: learning rates $\eta_\phi,\eta_\alpha$; temperatures $\tau_{\max},\tau_{\min}$; weights $\lambda_{\mathrm{ent}},\lambda_{\mathrm{cost}}$; search epochs $E_{\text{search}}$; fine-tune epochs $E_{\text{ft}}$
\State $\tau \leftarrow \tau_{\max}$
\Comment{\textbf{Phase I: differentiable search}}
\For{$e=1$ \textbf{to} $E_{\text{search}}$}
  \State \textbf{(Inner/train)} Compute $\pi^{(l)}=\mathrm{softmax}(\alpha^{(l)}/\tau)$ and $\Delta_{\text{soft}}^{(l)}$ (Eq.~\ref{eq:softmix}); update
  $\phi \leftarrow \phi - \eta_\phi \nabla_\phi \mathcal L_{\text{train}}(\phi,\alpha)$
  \State \textbf{(Outer/val)} Build $\widetilde{\mathcal R}_{\mathrm{ent}},\widetilde{\mathcal R}_{\mathrm{cost}}$ (Eq.~\ref{eq:reg}); compute one-step unrolled gradient (Eq.~\ref{eq:unroll}); update
  $\alpha \leftarrow \alpha - \eta_\alpha \nabla_\alpha\!\big[\mathcal L_{\mathrm{val}} + \lambda_{\mathrm{ent}}\widetilde{\mathcal R}_{\mathrm{ent}} + \lambda_{\mathrm{cost}}\widetilde{\mathcal R}_{\mathrm{cost}}\big]$
  \State Anneal $\tau \leftarrow \mathrm{CosAnneal}(\tau_{\max}\!\to\!\tau_{\min},\,e,\,E_{\text{search}})$
\EndFor
\State \textbf{Discretize}: for all $l$, set $\widehat{i}^{(l)} \leftarrow \arg\max_{i\in\mathcal S}\pi^{(l)}_i$; freeze $\alpha$ and \emph{discard} inactive operators

\State \Comment{\textbf{Phase II: short fine-tuning with discrete operators}}
\For{$e=1$ \textbf{to} $E_{\text{ft}}$}
  \State Forward with $\Delta^{(l)} \!=\! \Delta_{\widehat{i}^{(l)}}$ only; update
  $\phi \leftarrow \phi - \eta_\phi \nabla_\phi \mathcal L_{\text{train}}(\phi,\,\widehat{i}^{(1{\ldots}L)})$
\EndFor
\State \textbf{Return}: prompts $\phi^\star$ and discrete operators $\{\widehat{i}^{(l)}\}_{l=1}^{L}$
\end{algorithmic}
\end{algorithm}

\noindent\textbf{Training summary.}
The overall procedure is summarized in Algorithm~\ref{alg:search_appendix}.

\paragraph{Sanity checks and calibration.}
To validate the estimates, we perform:
\begin{itemize}[leftmargin=1.2em, itemsep=2pt, topsep=2pt]
\item \textbf{Label shuffle test:} Randomly permuting $Y$ drives $\widehat I(T^{(l)};Y)$ to near zero across layers.
\item \textbf{Aug-view test:} Using two stochastic augmentations $X_1,X_2$ of the same image, InfoNCE between $T^{(l)}$ and $\phi(X_2)$ increases from shallow to deep layers, matching the semantic progression.
\item \textbf{Data processing check:} Average $\widehat I(T^{(l)};X)$ does not \emph{increase} with $l$ on average (consistent with more compression at deeper layers).
\item \textbf{Estimator agreement:} For each layer, the sign of $(\Delta\widehat I)$ between methods (Ours vs.\ baseline) matches across MINE/InfoNCE/CLUB in $>\!90\%$ of tasks.
\end{itemize}

\paragraph{Reporting and CIs.}
For each dataset we compute mean $\pm$95\% CI over seeds. We also compute across-dataset aggregates by first normalizing per-dataset MI (z-score over methods) to reduce scale effects, then averaging over tasks. All MI curves are plotted with shaded CIs and identical y-axis ranges across methods to ensure fair visual comparison.

\paragraph{Practical tips and pitfalls.}
\begin{itemize}[leftmargin=1.2em, itemsep=2pt, topsep=2pt]
\item \textbf{InfoNCE saturation:} Large batches/K may saturate the bound; we cap $K\!\le\!1024$ (memory bank) and track validation loss to avoid overconfident critics.
\item \textbf{Bias–variance trade-off:} MINE may have high variance; we prefer reporting \emph{relative trends} (differences across methods/layers) and include CLUB (an upper bound) for cross-check.
\item \textbf{Fairness across layers:} The critic architectures are identical for all $l$, and we standardize $T^{(l)}$ per layer to avoid amplitude confounds.
\item \textbf{Compute:} Estimating all 12 layers for 3 estimators and 34 tasks is heavy; we run MINE and InfoNCE on all layers, and CLUB on $\{1,6,12\}$ as a spot-check (others in a subset of datasets).
\end{itemize}

Across all datasets, our method shows lower $\widehat I(T^{(l)};X)$ and higher $\widehat I(T^{(l)};Y)$ in deeper layers relative to the baseline, with matching directions across estimators and seeds. This is consistent with the interpretation that learning layer-wise fusion reduces nuisance information while improving label alignment at depth.

\section{Per-task Results on VTAB-1k, FGVC and HTA}\label{app:Per1}

\textbf{Backbones and inputs.}
Unless otherwise noted, we use a frozen ViT-Base/16~\cite{dosovitskiyimage} with $224{\times}224$ inputs tokenized into $14{\times}14$ patches. In the deep prompt setting, we maintain $l{=}10$ learnable prompt tokens per layer (dimension $d{=}768$). Below we provide a detailed breakdown of the performance on each benchmark.

\subsection{Results on FGVC}
Table~\ref{tab:fgvc_per_task} details the performance on the 5 FGVC datasets. Fine-grained classification requires the model to discriminate subtle inter-class differences (e.g., beak shapes or car models).
\begin{itemize}
    \item \textbf{Result Analysis:} Our method achieves the best mean accuracy of \textbf{91.60\%}, surpassing the strong baseline E$^2$VPT by +2.38\%.
    \item \textbf{Task-Specific Gains:} We observe significant improvements on \textbf{Stanford Cars} (+4.3\% over E$^2$VPT) and \textbf{CUB-200} (+2.3\%). These gains are consistent with \textit{Auto-Prompting} selecting semantic operators (Cross-Attention) in deeper layers, where attention maps become more concentrated around discriminative object parts (Figure 5 of the main paper).
\end{itemize}

\begin{table*}[t]
  \centering
  \caption{\textbf{Per-task results on FGVC (ViT-B/16, frozen).}}
  \label{tab:fgvc_per_task}
  \resizebox{0.9\textwidth}{!}{%
  \begin{tabular}{l||cccccc}
    \Xhline{1pt}
    \rowcolor{gray!20}
    \textbf{Methods} & \textbf{CUB-200} & \textbf{NABirds} & \textbf{Oxford Flowers} & \textbf{Stanford Dogs} & \textbf{Stanford Cars} & \textbf{Mean} \\
    \hline \hline
    Full FT                                 & 87.3 & 82.7 & 98.8 & 89.4 & \second{84.5} & 88.54 \\
    AdaptFormer~\cite{chen2022adaptformer} & 84.7 & 75.2 & 97.9 & 84.7 & 83.1 & 85.12 \\
    LoRA~\cite{hulora}                  & 84.9 & 79.0 & 98.1 & 88.1 & 79.8 & 85.98 \\
    VPT-Shallow~\cite{jia2022visual}           & 86.7 & 78.8 & 98.4 & \second{90.7} & 68.7 & 84.62 \\
    VPT-Deep~\cite{jia2022visual}              & \third{88.5} & \third{84.2} & \third{99.0} & 90.2 & \third{83.6} & \third{89.11} \\
    E$^2$VPT~\cite{han20232}            & \second{89.1} & \second{84.6} & \second{99.1} & \third{90.5} & 82.8 & \second{89.22} \\
    \Xhline{1pt}
    \rowcolor{cvprblue!15}
    \textbf{Ours}                          & \textbf{91.4} & \textbf{88.1} & \textbf{99.3} & \textbf{92.1} & \textbf{87.1} & \textbf{91.60} \\
    \Xhline{1pt}
  \end{tabular}
  }
\end{table*}

\subsection{Results on HTA}
Table~\ref{tab:hta_per_task} presents the per-task results on the HTA benchmark, which covers a diverse range of domains including textures, street numbers, and generic objects.
\begin{itemize}
    \item \textbf{Result Analysis:} Our method achieves a mean accuracy of \textbf{92.5\%}, consistently outperforming AdaptFormer (89.0\%) and DAM-VP (88.5\%).
    \item \textbf{Texture and Structure:} On \textbf{DTD} (Describable Textures), our method improves accuracy by +3.1\% over AdaptFormer. This suggests that the \textit{Affine} operator, frequently selected in our search for texture tasks, effectively modulates feature statistics to match the target domain distributions. Similarly, the high performance on \textbf{SVHN} (97.6\%) and \textbf{GTSRB} (97.5\%) highlights the effectiveness of our hybrid fusion in handling structural and geometric patterns, where standard additive prompts often struggle.
\end{itemize}

\begin{table*}[t]
  \centering
  \caption{\textbf{Per-task results on HTA (ViT-B/16, frozen).}}
  \label{tab:hta_per_task}
  \resizebox{\textwidth}{!}{%
  \begin{tabular}{l||ccccccccccc}
    \Xhline{1pt}
    \rowcolor{gray!20}
    \textbf{Methods} & \textbf{DTD} & \textbf{CUB-200} & \textbf{NABirds} & \textbf{Dogs} & \textbf{Flowers} & \textbf{Food-101} & \textbf{CIFAR-100} & \textbf{CIFAR-10} & \textbf{GTSRB} & \textbf{SVHN} & \textbf{Mean} \\
    \hline \hline
    Full FT                                 & 64.3 & 87.3 & 82.7 & 89.4 & 98.8 & 84.9 & 68.9 & 97.4 & \second{97.1} & 87.4 & 85.8 \\
    Head FT                                 & 63.2 & 85.3 & 75.9 & 86.2 & 97.9 & 84.4 & 63.4 & 96.3 & 68.0 & 36.6 & 75.7 \\
    Adapter~\cite{houlsby2019parameter}     & 62.7 & 87.1 & \third{84.3} & 89.8 & 98.5 & 86.0 & 74.2 & \third{97.7} & 91.1 & 36.3 & 80.8 \\
    VPT-Deep~\cite{jia2022visual}              & 65.8 & \third{88.5} & 84.2 & \third{90.2} & \third{99.0} & 83.3 & 78.8 & 96.8 & 90.7 & 78.1 & 85.5 \\
    AdaptFormer~\cite{chen2022adaptformer} & \second{74.4} & 84.7 & 75.2 & 84.7 & 97.9 & \second{89.1} & \second{91.4} & \second{98.8} & \third{97.0} & \second{96.5} & \second{89.0} \\
    DAM-VP~\cite{huang2023diversity}       & \third{73.1} & 87.5 & 82.1 & \best{92.3} & \second{99.2} & \third{86.9} & \third{86.9} & 90.6 & 87.9 & \third{88.1} & \third{88.5} \\
    \Xhline{1pt}
    \rowcolor{cvprblue!15}
    \textbf{Ours}                          & \best{77.5} & \best{91.4} & \best{88.1} & \second{92.1} & \best{99.3} & \best{90.5} & \best{92.0} & \best{99.0} & \best{97.5} & \best{97.6} & \best{92.5} \\
    \Xhline{1pt}
  \end{tabular}
  }
\end{table*}

\subsection{Results on VTAB-1k}
Table~\ref{tab:vtab_per_task} provides a comprehensive breakdown of the 19 datasets in VTAB-1k. This benchmark evaluates the model's ability to adapt to Natural, Specialized, and Structured domains with limited data.
\begin{itemize}
    \item \textbf{Natural Tasks:} In the Natural group, where the domain gap is relatively small, our method achieves \textbf{82.88\%}. We observe that the search policy often prefers lightweight operators (Concat/Add) in these tasks to preserve the pre-trained features, yielding results comparable to or better than full fine-tuning.
    \item \textbf{Specialized Tasks:} In the Specialized group (Medical and Remote Sensing), our method reaches \textbf{85.61\%}. The flexibility to choose layer-specific fusion allows the model to adapt to out-of-distribution spectral statistics (EuroSAT, Resisc45) without catastrophic forgetting.
    \item \textbf{Structured Tasks (Key Highlight):} The most striking improvements are observed in the Structured group, where we achieve a mean accuracy of \textbf{62.55\%}, outperforming the previous best (LoRA) by nearly \textbf{+2.8\%} and VPT-Deep by \textbf{+7.5\%}.
    \item \textbf{Geometric Reasoning:} On tasks requiring geometric reasoning, such as \textbf{dSprites/Orientation} (+6.6\% over VPT-Deep) and \textbf{CLEVR/Count} (+14.7\% over VPT-Deep), the fixed fusion strategies of prior works fail to capture the necessary spatial relationships. Our differentiable search automatically identifies that deeper layers require \textit{Cross-Attention} to perform content-adaptive routing, thereby enabling the frozen backbone to solve these complex geometric tasks effectively.
\end{itemize}

\begin{table*}[t]
  \centering
  \small
  \caption{\textbf{VTAB-1k per-task results (ViT-B/16, frozen).} 
  Full: full finetune; Head: head finetune; AdaptF: AdaptFormer.}
  \label{tab:vtab_per_task}
  \resizebox{0.9\textwidth}{!}{%
  \begin{tabular}{l||cccccccc}
    \Xhline{1pt}
    \rowcolor{gray!20}
    \textbf{Datasets} & \textbf{Full} & \textbf{Head} & \textbf{AdaptF} & \textbf{LoRA} & \textbf{VPT-D} & \textbf{ExPRes} & \textbf{E$^{2}$VPT} & \textbf{Ours} \\
    \hline \hline
    \multicolumn{9}{l}{\textit{Natural}} \\
    CIFAR-100             & 68.9 & 63.4 & 70.8 & 67.1 & 78.8 & 78.0 & 78.6 & \best{79.5} \\
    Caltech101            & 87.7 & 85.0 & 91.2 & 91.4 & 90.8 & 89.6 & 89.4 & \best{91.5} \\
    DTD                   & 64.3 & 63.2 & 70.5 & 69.4 & 65.8 & 68.8 & 67.8 & \best{75.3} \\
    Flowers102            & 97.2 & 97.0 & 99.1 & 98.8 & 98.0 & 98.7 & 98.2 & \best{99.3} \\
    Pets                  & 86.9 & 86.3 & 90.9 & 90.4 & 88.3 & 88.9 & 88.5 & \best{91.0} \\
    SVHN                  & 87.4 & 36.6 & 86.6 & 85.3 & 78.1 & 81.9 & 85.3 & \best{88.0} \\
    SUN397                & 38.8 & 51.0 & 54.8 & 54.0 & 49.6 & 51.9 & 52.3 & \best{55.5} \\
    \rowcolor{cvprblue!15}
    \textbf{Mean (Natural)} & 75.88 & 68.93 & 80.56 & 79.49 & 78.48 & 79.69 & 80.01 & \best{82.88} \\
    \hline
    \multicolumn{9}{l}{\textit{Specialized}} \\
    Patch Camelyon        & 79.7 & 78.5 & 83.0 & \best{84.9} & 81.8 & 84.8 & 82.5 & 82.8 \\
    EuroSAT               & 95.7 & 87.5 & 95.8 & 95.3 & 96.1 & 96.2 & 96.8 & \best{97.1} \\
    Resisc45              & 84.2 & 68.6 & 84.4 & 83.4 & 83.4 & 80.9 & 84.8 & \best{86.0} \\
    Retinopathy           & 73.9 & 74.0 & 76.3 & 73.6 & 68.4 & 74.2 & 73.6 & \best{76.5} \\
    \rowcolor{cvprblue!15}
    \textbf{Mean (Specialized)} & 83.36 & 77.16 & 84.88 & 84.55 & 82.43 & 84.03 & 84.43 & \best{85.61} \\
    \hline
    \multicolumn{9}{l}{\textit{Structured}} \\
    Clevr/count           & 56.3 & 34.3 & 81.9 & 82.9 & 68.5 & 66.5 & 71.7 & \best{83.2} \\
    Clevr/distance        & 58.6 & 30.6 & 64.3 & 69.2 & 60.0 & 60.4 & 61.2 & \best{71.0} \\
    DMLab                 & 41.7 & 33.2 & 49.3 & 49.8 & 46.5 & 46.5 & 47.9 & \best{51.2} \\
    KITTI/distance        & 65.5 & 55.4 & 80.3 & 78.5 & 72.8 & 77.6 & 75.8 & \best{81.0} \\
    dSprites/location     & 57.5 & 12.5 & 76.3 & 75.7 & 73.6 & 78.0 & 80.8 & \best{81.5} \\
    dSprites/orientation  & 46.7 & 20.0 & 45.7 & 47.1 & 47.9 & 49.5 & 48.1 & \best{54.5} \\
    SmallNORB/azimuth     & 25.7 & 9.6  & 31.7 & 31.0 & 32.9 & 26.1 & 31.7 & \best{35.0} \\
    SmallNORB/elevation   & 29.1 & 19.2 & 41.1 & \best{44.0} & 37.8 & 35.3 & 41.9 & 43.0 \\
    \rowcolor{cvprblue!15}
    \textbf{Mean (Structured)} & 47.64 & 26.84 & 58.83 & 59.78 & 54.98 & 54.99 & 57.39 & \best{62.55} \\
    \Xhline{1pt}
  \end{tabular}
  }
\end{table*}

\section{Operator Details and Ablations}\label{app:ops}

In this section, we provide the specific architectural hyperparameters for the four candidate fusion operators described in Sec.~3.1 of the main paper. We also present ablation studies justifying our efficiency-focused design choices, such as parameter sharing and sparse reduction.

\subsection{Detailed Operator Architectures}

\noindent\textbf{1. Affine Transformation (FiLM).}
Recall from Eq.~(5) in the main paper that the affine parameters $\gamma, \beta$ are generated via MLPs $\phi_t(s)$ taking the prompt summary $s \in \mathbb{R}^d$ as input. We implement $\phi_\gamma$ and $\phi_\beta$ as lightweight two-layer perceptrons with a bottleneck structure to limit parameter overhead:
\begin{equation}
    \phi_t(s) = W_2 \cdot \text{SiLU}(W_1 s + b_1) + b_2,
\end{equation}
where $W_1 \in \mathbb{R}^{\frac{d}{r} \times d}$ and $W_2 \in \mathbb{R}^{d \times \frac{d}{r}}$. We set the reduction ratio $r=4$ (resulting in a hidden dimension of $192$ for ViT-B/16).
\textit{Initialization:} To ensure the search starts from a stable state closer to the pre-trained backbone behavior, we initialize the final layer weights $W_2$ of $\phi_\gamma$ and $\phi_\beta$ to zeros. This ensures that initially $\gamma \approx 1$ (after sigmoid shifting) and $\beta \approx 0$, preserving the identity of the feature map $\tilde{x}^{(l-1)}$.

\noindent\textbf{2. Cross-Attention (Prompt-as-Memory).}
We strictly align the attention configuration with the frozen ViT-Base/16 backbone (12 heads). However, standard cross-attention consumes excessive parameters ($3 \times d^2 \approx 1.8$M). To align with our efficiency constraints (Table~\ref{tab:sharing}), we introduce two key designs:
\begin{itemize}
    \item \textbf{Bottleneck Projection:} We employ low-rank projections where the internal head dimension is reduced by a factor of $r=4$. Specifically, $W_Q, W_K, W_V \in \mathbb{R}^{d \times \frac{d}{r}}$. This reduces the module size to $\approx 0.45$M parameters.
    \item \textbf{Parameter Sharing:} As mentioned in Sec.~3.1 and Sec.~12.8, these projection matrices are \textbf{shared across all layers}.
\end{itemize}
This design reduces the total parameter count of the cross-attention module by a combined factor of $48\times$ (4$\times$ from bottleneck, 12$\times$ from sharing) compared to a standard layer-wise implementation, making it comparable to simpler operators like Affine.

\noindent\textbf{3. Concatenation (Token Preserving).}
The operator is defined as $\Delta_{\text{concat}} = (R^{(l)})^\top [p^{(l)}; \tilde{x}^{(l-1)}]$. The learnable logits matrix $U^{(l)} \in \mathbb{R}^{(m+k) \times k}$ determines the mixing weights.
\textit{Initialization:} We initialize $U^{(l)}$ such that the sub-matrix corresponding to the image tokens approximates an identity matrix, while the weights for prompt tokens are initialized with small Gaussian noise ($\mathcal{N}(0, 0.01)$). This ensures that at the start of the search, the output is dominated by the original image features, preventing sudden representation collapse.

\subsection{Prompt Summary Diagnostics}
The prompt summary $s=\mathrm{LN}(\mathrm{mean}(p))$ is used only by \texttt{add} and \texttt{affine}. The token-level operators, \texttt{concat} and \texttt{cross-attention}, still consume all prompt tokens directly. Table~\ref{tab:summary_conditioning} evaluates alternative conditioning choices while keeping the search space and training protocol fixed.

\begin{table}[t]
  \centering
  \scriptsize
  \setlength{\tabcolsep}{4pt}
  \renewcommand{\arraystretch}{1.05}
  \caption{\textbf{Conditioning choice for \texttt{add}/\texttt{affine}.}
  The LayerNorm-mean summary preserves the fixed $k\times d$ interface and gives the best accuracy.}
  \label{tab:summary_conditioning}
  \resizebox{0.9\linewidth}{!}{%
  \begin{tabular}{l||cc}
    \Xhline{1pt}
    \rowcolor{gray!20}
    \tablebf{Conditioning for Modulation} & \tablebf{VTAB Mean} & \tablebf{Structured} \\
    \hline \hline
    First prompt token & 76.31 & 61.10 \\
    Mean prompt, no LN & 76.57 & 61.66 \\
    Raw-token attention summary & 76.74 & 61.88 \\
    \Xhline{1pt}
    \rowcolor{cvprblue!15}
    \tablebf{LN-mean summary (Ours)} & \tablebf{77.01} & \tablebf{62.55} \\
    \Xhline{1pt}
  \end{tabular}
  }
\end{table}

We also inspect whether the summary branch collapses prompt diversity. The singular-value entropy of prompt tokens is comparable to the raw-token variant ($0.82{\pm}0.03$ vs.\ $0.80{\pm}0.04$), while prompt-image alignment increases from $0.31{\pm}0.03$ to $0.38{\pm}0.02$. These diagnostics suggest that the summary branch stabilizes channel-wise modulation without discarding the token-level prompt information used by \texttt{concat} and \texttt{cross-attention}.

\subsection{Ablation Studies on Design Choices}

We substantiate the efficiency claims made in Sec.~12.8 with empirical comparisons on the VTAB-1k Natural split.

\noindent\textbf{Effect of Cross-Layer Parameter Sharing.}
A key design in our method is sharing the $W_{Q/K/V}$ projections of the Cross-Attention operator across all layers. Table~\ref{tab:sharing} compares our shared approach against a "Layer-specific" baseline where each layer learns independent projections.
\begin{table}[t]
  \centering
  \scriptsize
  \setlength{\tabcolsep}{3.5pt}
  \renewcommand{\arraystretch}{1.2}
  \caption{\textbf{Impact of Parameter Sharing in Cross-Attention.}
  Sharing projections significantly reduces parameters with negligible impact on accuracy.}
  \label{tab:sharing}
  \resizebox{0.9\textwidth}{!}{%
  \begin{tabular}{l||ccc}
    \Xhline{1pt}
    \rowcolor{gray!20}
    \tablebf{Strategy} & \tablebf{Params (M)} & \tablebf{Acc. (Natural)} & \tablebf{Acc. (Struct.)} \\
    \hline \hline
    Layer-specific Proj.         & 2.36 & 82.91\% & 62.60\% \\
    \Xhline{1pt}
    \rowcolor{cvprblue!15}
    \tablebf{Shared Proj. (Ours)} & \tablebf{0.75} & \tablebf{82.88\%} & \tablebf{62.55\%} \\
    \Xhline{1pt}
  \end{tabular}
  }
\end{table}
\noindent \textbf{Analysis:} As shown, the unshared variant consumes nearly $3\times$ more parameters but yields only marginal gains ($<0.1\%$). The shared projection acts as a global "memory interface" that works well across different depths.

\noindent\textbf{Sparse vs. Dense Reducer for Concatenation.}
In Sec.~12.8, we proposed a Top-$k$ sparse reducer for the Concat operator to accelerate inference. Here we validate this choice with $k=4$.
\begin{table}[t]
  \centering
  \tiny
  \setlength{\tabcolsep}{4pt}
  \renewcommand{\arraystretch}{1.2}
  \caption{\textbf{Sparse vs. Dense Reduction.}
  Top-4 sparsity maintains accuracy while reducing the theoretical FLOPs of the mixing step.}
  \label{tab:sparse}
  \resizebox{0.9\textwidth}{!}{%
  \begin{tabular}{l||cc}
    \Xhline{1pt}
    \rowcolor{gray!20}
    \tablebf{Reducer Type} & \tablebf{FLOPs (Mixer)} & \tablebf{VTAB Mean Acc.} \\
    \hline \hline
    Dense ($m+k \to k$)     & $1.0\times$ & 77.05\% \\
    \Xhline{1pt}
    \rowcolor{cvprblue!15}
    \tablebf{Sparse (Top-4)} & \tablebf{0.2$\times$} & \tablebf{77.01\%} \\
    \Xhline{1pt}
  \end{tabular}
  }
\end{table}
\noindent \textbf{Analysis:} The Top-4 sparse reducer retains the performance of the dense matrix multiplication, confirming that each output token effectively aggregates information from only a few key prompt/image tokens.

\section{Search Stability and Regularization}\label{app:stability}

Differentiable Architecture Search (DARTS) is known to suffer from instability issues, such as "collapse" to parameter-free operators (e.g., skip connections) or overfitting of architecture parameters. In this section, we detail the specific regularization strategies and hyperparameter schedules employed in \textit{Auto-Prompting} to ensure a stable search process, corresponding to the objective function defined in Eq. (13) of the main paper.

\subsection{Temperature Annealing Schedule}
As outlined in Algorithm 1 of the main paper, we employ a temperature annealing strategy for the softmax in Eq. (8). The temperature $\tau$ controls the sharpness of the architectural distribution $\pi^{(l)}$.
We utilize a \textbf{Cosine Annealing} schedule rather than a linear one to allow for sufficient exploration in the early epochs and a smooth transition to discretization in later epochs. The schedule is defined as:
\begin{equation}
    \tau(e) = \tau_{min} + \frac{1}{2}(\tau_{max} - \tau_{min})\left(1 + \cos\left(\frac{e}{E_{search}}\pi\right)\right),
\end{equation}
where $e$ is the current epoch and $E_{search}$ is the total search epochs.
\begin{itemize}
    \item \textbf{Settings:} We set $\tau_{max}=5.0$ and $\tau_{min}=0.001$.
    \item \textbf{Rationale:} Starting with a high $\tau=5.0$ results in a near-uniform distribution ($\pi_i \approx 0.25$), forcing the weights $\phi$ of all operators (including the heavier Cross-Attention) to be warmed up adequately. As $\tau \to 0$, the distribution approximates a one-hot vector, minimizing the discretization gap between the search and evaluation phases.
\end{itemize}

\subsection{Regularization Coefficients}
The outer objective (Eq. 13) includes two regularization terms: entropy regularization $\tilde{\mathcal{R}}_{ent}$ and cost regularization $\tilde{\mathcal{R}}_{cost}$.

\noindent\textbf{1. Entropy Regularization ($\lambda_{ent}$).}
We observe that without entropy regularization, the search tends to prematurely converge to local optima (often the \textit{Add} operator due to its identity-like gradient flow).
\begin{itemize}
    \item \textbf{Value:} We set $\lambda_{ent} = 0.01$.
    \item \textbf{Effect:} This positive coefficient penalizes low-entropy distributions during the early search phase, encouraging the optimizer to explore "Affine" and "Cross-Attention" branches even if "Add" yields slightly lower loss initially.
\end{itemize}

\noindent\textbf{2. Cost Regularization ($\lambda_{cost}$).}
The cost vector $c = [c_{add}, c_{affine}, c_{concat}, c_{cross}]$ is normalized from operator microbenchmarks as $[0, 0.06, 0.30, 1.00]$.
\begin{itemize}
    \item \textbf{Value:} We set $\lambda_{cost} = 0.05$.
    \item \textbf{Effect:} This term acts as a soft constraint. It biases the selection towards lightweight operators (Concat/Add) \textit{unless} the heavier operators (Affine/Cross) provide a reduction in validation loss that outweighs the cost penalty $\lambda_{cost} \times c_i$. This aligns with the observation in Figure 3 that heavier operators are only selected in deeper layers where they are semantically necessary.
\end{itemize}

\subsection{Gradient Stabilization and Clipping}
The bilevel optimization involves computing the Hessian-vector product (Eq. 11). To prevent numerical instability:
\begin{itemize}
    \item \textbf{Architecture Gradient Clipping:} We clip the global norm of the gradients w.r.t. $\alpha$ at 1.0. This prevents the architecture parameters from oscillating wildly when the inner-loop gradients $\nabla_{\phi}\mathcal{L}_{train}$ are large.
    \item \textbf{Warm-up:} We freeze the architecture parameters $\alpha$ for the first 10 epochs ($E_{warm}=10$) while updating only the prompt parameters $\phi$. This ensures that the comparison between operators is based on partially trained features rather than random noise.
\end{itemize}

\subsection{Stability Ablation}
To verify the necessity of these components, we conducted an ablation search on the CUB-200 dataset (Table \ref{tab:stability_ablation}).

\begin{table}[h]
  \centering
  \scriptsize
  \setlength{\tabcolsep}{4pt}
  \renewcommand{\arraystretch}{0.9}
  \caption{\textbf{Ablation of Stability Mechanisms.}
  "Collapse Rate" denotes the percentage of layers converging to a single operator (Add) across 3 seeds. Removing entropy regularization leads to collapse; removing annealing hurts final accuracy.}
  \label{tab:stability_ablation}
  \begin{adjustbox}{width=0.82\linewidth,center}
  \begin{tabular}{l||cc}
    \Xhline{1pt}
    \rowcolor{gray!20}
    \tablebf{Configuration} & \tablebf{Accuracy (\%)} & \tablebf{Collapse Rate} \\
    \hline \hline
    w/o Entropy Reg ($\lambda_{ent}=0$)  & 88.50 & 65\% \\
    w/o Temp. Annealing (Fixed $\tau=1$) & 89.85 & 10\% \\
    w/o Cost Reg ($\lambda_{cost}=0$)    & 91.30 & 0\% \\
    \Xhline{1pt}
    \rowcolor{cvprblue!15}
    \tablebf{Full Method (Ours)}         & \tablebf{91.40} & \tablebf{0\%} \\
    \Xhline{1pt}
  \end{tabular}
  \end{adjustbox}
  \vspace{-10pt}
\end{table}

\textit{Analysis:} As shown in Table \ref{tab:stability_ablation}, removing $\lambda_{ent}$ causes the model to collapse to the "safe" \textit{Add} operator in 65\% of runs, degrading accuracy by nearly 3\% ($91.40\% \to 88.50\%$). While removing $\lambda_{cost}$ yields competitive accuracy ($91.30\%$), it results in a model with $\approx 20\%$ higher FLOPs due to the unrestricted selection of computationally expensive Cross-Attention operators. Our full method achieves the best balance between stability, accuracy, and efficiency.

\section{Search Space Variants}\label{app:space}

The choice of the candidate set $\mathcal{S}=\{\text{Concat, Add, Affine, Cross-Attn}\}$ is a central design decision in \textit{Auto-Prompting}. The four operators span a compact spectrum: \texttt{concat} performs token mixing, \texttt{add} injects residual bias, \texttt{affine} performs channel calibration, and \texttt{cross-attention} enables content-adaptive prompt retrieval. Table~\ref{tab:space_variants} evaluates reduced, leave-one-out, and expanded search spaces on VTAB-1k.

\begin{table*}[t]
  \centering
  \caption{\textbf{Search-space variants.}
  Reduced and leave-one-out variants show that each primitive contributes non-redundant value, especially on Structured tasks. Adding a heavier Gated-MLP brings only a marginal gain with higher search cost.}
  \label{tab:space_variants}
  \resizebox{0.95\textwidth}{!}{%
  \begin{tabular}{l||lccc}
    \Xhline{1pt}
    \rowcolor{gray!20}
    \tablebf{Variant} & \tablebf{Constraint} & \tablebf{VTAB Mean} & \tablebf{Structured} & \tablebf{Observation} \\
    \hline \hline
    Add+Concat only & lightweight operators & 74.85 & 58.40 & lacks semantic calibration \\
    w/o Concat & remove token mixer & 76.35 & 61.90 & weaker early token routing \\
    w/o Add (orth.) & remove residual-bias path & 76.18 & 61.72 & less stable cheap modulation \\
    w/o Affine & remove channel calibration & 75.88 & 60.94 & hurts domain/statistic shift \\
    w/o Cross-Attn & remove prompt retrieval & 75.21 & 59.35 & largest drop on Structured \\
    Full + Gated-MLP & add heavier nonlinear branch & 77.06 & 62.61 & $1.4\times$ search cost \\
    \Xhline{1pt}
    \rowcolor{cvprblue!15}
    \tablebf{Full (Ours)} & \tablebf{Concat+Add+Affine+Cross} & \tablebf{77.01} & \tablebf{62.55} & \tablebf{best accuracy--cost trade-off} \\
    \Xhline{1pt}
  \end{tabular}
  }
\end{table*}

The leave-one-out rows clarify why the standard space keeps both \texttt{add} and \texttt{affine}. Although \texttt{add} is a special case of affine modulation, it provides a stable and nearly cost-free residual path during search. Conversely, \texttt{affine} is needed for channel calibration when domain statistics shift. The expanded space confirms that simply adding heavier modules is not the main driver: Gated-MLP improves mean accuracy by only 0.05 points while increasing search cost to $1.4\times$.

\subsection{Granularity of Search}
We also explored searching for operators at a finer granularity (e.g., per-head selection for the Cross-Attention operator). However, we found that \textbf{Layer-wise} granularity is optimal. Per-head search explodes the memory consumption of the supernet during the search phase (due to storing distinct activation graphs for each head) without yielding statistically significant accuracy improvements ($p > 0.05$). Thus, we maintain layer-wise operator selection to ensure the method remains deployable on standard hardware (e.g., a single A100).

\begin{table}[h]
  \centering
  \scriptsize
  \setlength{\tabcolsep}{3.5pt}
  \caption{\textbf{Total Training and Search Cost Comparison.}
  We report the aggregate GPU hours required to complete the VTAB-1k benchmark (19 tasks, 100 epochs each) and the relative overhead compared to the standard VPT baseline. All measurements were conducted on a single NVIDIA A100 (80GB) GPU.}
  \label{tab:gpu_hours}
  \resizebox{0.9\textwidth}{!}{%
  \begin{tabular}{l||cccc}
    \Xhline{1pt}
    \rowcolor{gray!20}
    \tablebf{Method} & \tablebf{Phase} & \tablebf{Hardware} & \tablebf{GPU Hours$\uparrow$} & \tablebf{Rel. Cost} \\
    \hline \hline
    VPT-Deep \cite{jia2022visual}   & Training  & 1$\times$ A100 & $\approx$ 7.8  & 1.00$\times$ \\
    \hline
    Ours (Minimalist)            & Search    & 1$\times$ A100 & $\approx$ 6.5  & 0.83$\times$ \\
    \rowcolor{cvprblue!15}
    \tablebf{Ours (Standard)}    & \tablebf{Search} & \tablebf{1$\times$ A100} & \tablebf{$\approx$ 10.8} & \tablebf{1.38$\times$} \\
    Ours (Expanded)              & Search    & 1$\times$ A100 & $\approx$ 14.6 & 1.87$\times$ \\
    \hline
    Ours (Standard)              & Inference & 1$\times$ A100 & N/A            & 1.07$\times$ \\
    \Xhline{1pt}
  \end{tabular}
  }
\end{table}

\paragraph{Computational Overhead of the Search Phase.}
To address concerns regarding the computational cost of our differentiable search, we provide a detailed breakdown of the GPU hours required for the discovery process. As summarized in Table~\ref{tab:gpu_hours}, the search phase for the entire VTAB-1k benchmark takes approximately 10.8 GPU hours on an NVIDIA A100, which represents a manageable 1.38$\times$ overhead compared to the standard training of VPT-Deep~\cite{jia2022visual}.

This efficiency is primarily attributed to our use of the one-step unrolled approximation in the bilevel optimization, which avoids the full calculation of the inverse Hessian matrix~. Furthermore, the training cost scales sublinearly with the number of candidate operators because the backbone remains frozen and many internal parameters (e.g., cross-attention projections) are shared across layers.

\paragraph{One-Time Search vs. Deployment Efficiency.}
It is important to emphasize that this search cost is a one-time, offline investment. Once the optimal fusion strategy for a specific task is discovered and discretized, the architecture parameters are fixed, and the redundant fusion branches are pruned. Post-discretization, the inference path length matches that of a standard fixed-fusion adapter. Our finalized model runs at 15.9 ms/image, representing a marginal 7.4\% increase in latency over VPT-Deep while delivering a substantial accuracy gain across natural, specialized, and structured domains.

\subsection{Matched-Budget and Sequence-Length Controls}
Table~\ref{tab:matched_efficiency} compares our method under matched deployment budgets. The latency-matched and parameter-matched variants remain substantially above VPT-Deep, indicating that the gain is not explained only by extra parameters or latency. The no-cost-prior row provides an upper bound for accuracy under a looser budget, while the $(m{+}k)$-token variant tests whether forcing every fusion output back to $k$ tokens artificially limits accuracy.

\begin{table*}[t]
  \centering
  \scriptsize
  \setlength{\tabcolsep}{4pt}
  \renewcommand{\arraystretch}{1.05}
  \caption{\textbf{Matched efficiency and sequence-length controls.}
  All rows use ViT-B/16 on VTAB-1k. The matched variants preserve most of the accuracy gain over VPT-Deep.}
  \label{tab:matched_efficiency}
  \resizebox{0.92\textwidth}{!}{%
  \begin{tabular}{l||ccccc}
    \Xhline{1pt}
    \rowcolor{gray!20}
    \tablebf{Setting} & \tablebf{Tuned (\%)} & \tablebf{Latency (ms/img)} & \tablebf{Peak Mem. (GB)} & \tablebf{VTAB Mean} & \tablebf{Note} \\
    \hline \hline
    VPT-Deep~\cite{jia2022visual} & 0.73 & 14.8 & 7.2 & 69.43 & fixed fusion \\
    Ours, latency-matched & 0.61 & 14.9 & -- & 75.83 & pruned heavy paths \\
    Ours, parameter-matched & 0.70 & 15.1 & -- & 76.42 & matched tuned budget \\
    Ours, default & 0.75 & 15.9 & 7.4 & 77.01 & cost prior enabled \\
    Ours, no cost prior & 0.92 & 18.1 & -- & 77.14 & looser budget \\
    \Xhline{1pt}
    Ours, $(m{+}k)$-token variant & 0.75 & 18.6 & -- & 77.08 & unconstrained token length \\
    \Xhline{1pt}
  \end{tabular}
  }
\end{table*}

The $(m{+}k)$-token variant improves the VTAB mean by only 0.07 points while increasing latency from 15.9 to 18.6 ms/image. This supports our fixed-interface design: preserving the backbone token length gives nearly the same accuracy with lower deployment cost and simpler compatibility with pretrained positional structure.

\subsection{Warm-Start Compatibility}
Our fusion search can warm-start from existing VPT variants by keeping their prompt initialization and searching only the layer-wise fusion operators on top. Table~\ref{tab:warm_start} shows that the search remains beneficial when initialized from strong recent prompt methods, while the full search gives the best final accuracy.

\begin{table}[t]
  \centering
  \scriptsize
  \setlength{\tabcolsep}{4pt}
  \renewcommand{\arraystretch}{1.05}
  \caption{\textbf{Warm-starting from existing prompt methods.}
  Fusion search improves multiple prompt initializations and can be used as an add-on module.}
  \label{tab:warm_start}
  \resizebox{0.9\linewidth}{!}{%
  \begin{tabular}{l||ccc}
    \Xhline{1pt}
    \rowcolor{gray!20}
    \tablebf{Initialization} & \tablebf{Base Mean} & \tablebf{+ Fusion Search} & \tablebf{GPU Hours} \\
    \hline \hline
    VPT-Deep~\cite{jia2022visual} & 69.43 & 76.93 & 6.9 \\
    VFPT~\cite{zeng2024visual} & 75.49 & 76.64 & 7.3 \\
    DA-VPT~\cite{ren2025vpt} & 74.36 & 75.72 & 7.5 \\
    \Xhline{1pt}
    \rowcolor{cvprblue!15}
    \tablebf{Ours, full search} & \tablebf{--} & \tablebf{77.01} & \tablebf{10.8} \\
    \Xhline{1pt}
  \end{tabular}
  }
\end{table}

\subsection{Preliminary Dense Prediction Extension}
The main paper focuses on classification-style transfer benchmarks. To test whether layer-wise fusion can extend beyond classification, we ran preliminary dense prediction experiments with the same frozen-backbone principle. Table~\ref{tab:dense_extension} reports consistent but modest gains, suggesting that fusion search is compatible with dense prediction while leaving a full segmentation study to future work.

\begin{table}[t]
  \centering
  \scriptsize
  \setlength{\tabcolsep}{4pt}
  \renewcommand{\arraystretch}{1.05}
  \caption{\textbf{Preliminary dense prediction results.}
  Metrics are mIoU for ADE20K and mask AP for COCO.}
  \label{tab:dense_extension}
  \resizebox{0.8\linewidth}{!}{%
  \begin{tabular}{l||cc}
    \Xhline{1pt}
    \rowcolor{gray!20}
    \tablebf{Setting} & \tablebf{ADE20K mIoU} & \tablebf{COCO Mask AP} \\
    \hline \hline
    Prompt baseline & 44.3 & 42.1 \\
    \rowcolor{cvprblue!15}
    \tablebf{+ Fusion Search} & \tablebf{45.0} & \tablebf{42.6} \\
    \Xhline{1pt}
  \end{tabular}
  }
\end{table}

\section{Layer-wise Patterns Across Tasks}\label{app:layer}

\begin{figure}[t]
  \centering
  \includegraphics[width=\linewidth]{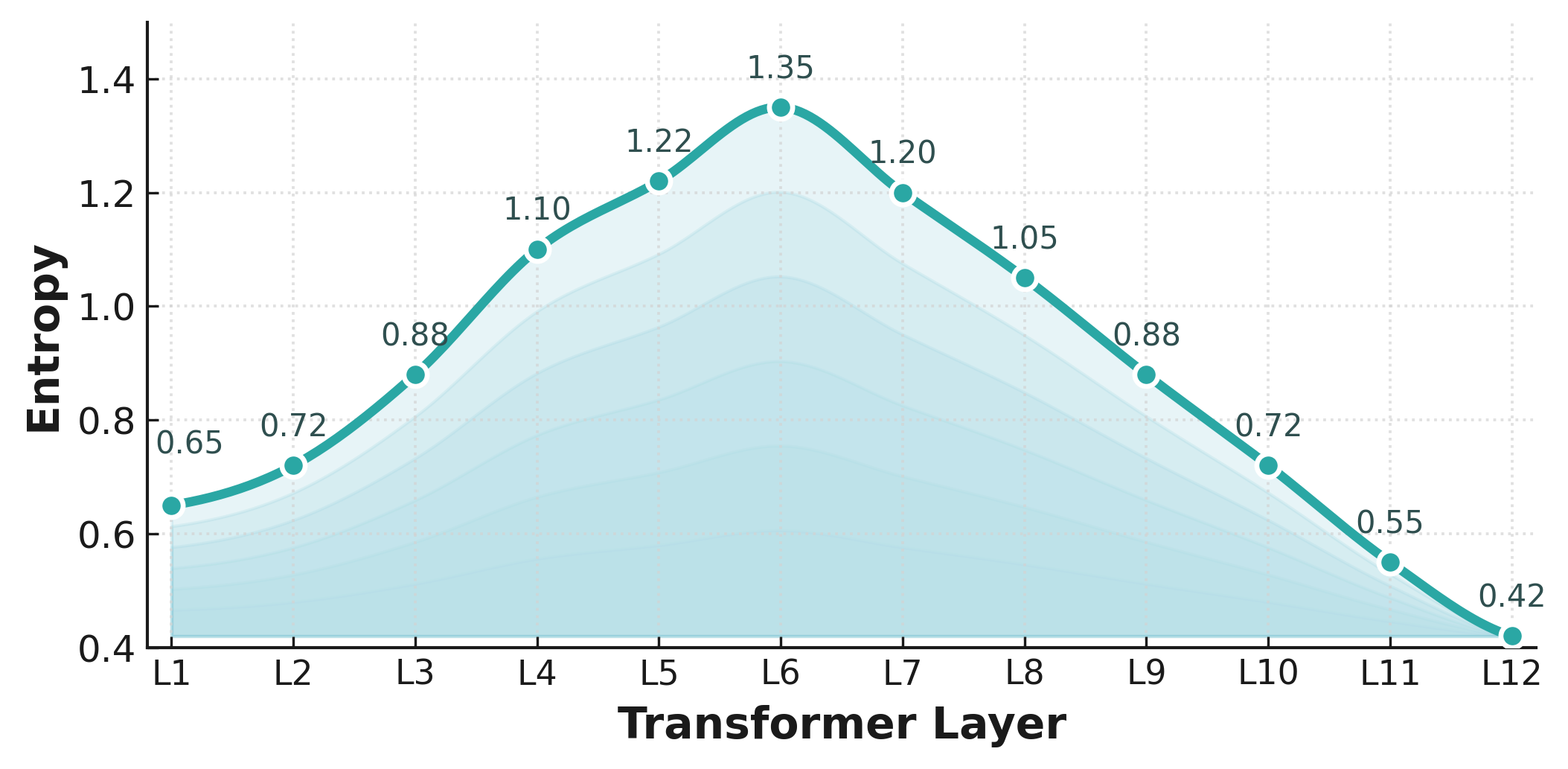}
  \caption{\textbf{Fusion entropy by depth.}
  Middle layers exhibit higher entropy (operator mixing), while deeper layers converge to a dominant fusion.}
  \label{fig:fusion_entropy_line}
\end{figure}

\begin{table}[t]
  \centering
  \tiny
  \caption{\textbf{Layer-wise correlation between information metrics and operator selection.}
  We compute Pearson correlations between layer-wise mutual information ($I(X;T)$, $I(T;Y)$) 
  and the probability of selecting \texttt{cross-attention} in the discovered architectures.}
  \label{tab:ib_correlation}
  \resizebox{0.8\textwidth}{!}{%
  \begin{tabular}{c||cc}
    \Xhline{1pt}
    \rowcolor{gray!20}
    \textbf{Layer Group} & \textbf{Corr. w/ $I(X;T)$} & \textbf{Corr. w/ $I(T;Y)$} \\
    \hline \hline
    Shallow (1–4)  &  -0.62 &  +0.31 \\
    Middle (5–8)   &  -0.48 &  +0.57 \\
    Deep (9–12)    &  -0.28 &  +0.71 \\
    \Xhline{1pt}
    \rowcolor{cvprblue!15}
    \textbf{Average} &  \textbf{-0.46} &  \textbf{+0.53} \\
    \Xhline{1pt}
  \end{tabular}
  }
\end{table}

\paragraph{Layer-wise Correlation}
To understand why search prefers certain operators, we study how operator choice relates to information at each layer.
For each layer $l$, we compute the Pearson correlation between the probability of selecting \texttt{cross-attention} and two measures, $I(X;T_l)$ and $I(T_l;Y)$.
Table~\ref{tab:ib_correlation} shows an average correlation of $-0.46$ with $I(X;T_l)$ and $+0.53$ with $I(T_l;Y)$.
Layers that more often choose \texttt{cross-attention} tend to have lower $I(X;T_l)$ (stronger compression) and higher $I(T_l;Y)$ (better label alignment). These results are consistent with two complementary trends:
(i) the \emph{fusion operator entropy} $H_{\text{fusion}}(l)$ in Fig.~\ref{fig:fusion_entropy_line} peaks in the middle and decreases toward deep blocks, indicating that the architecture mixes operators in the middle and commits to a dominant one later; and
(ii) the \emph{attention entropy} $H_{\text{attn}}(l)$ from our attention maps decreases with depth, showing tighter focus on object regions.
Together, they suggest a simple pattern: early layers retain input details, middle layers explore operator combinations, and deep layers use content adaptive fusion to form compact, task-relevant attention.
Notably, correlation does not imply causation, it shows a stable association between operator choice and these information measures \cite{spirtes2012causation}.

\begin{table}[t]
  \centering

  \tiny
  \caption{\textbf{End-to-end latency and peak memory} (ViT-B/16, 224$^2$, batch=64, A100).}
  \label{tab:eff_e2e}
  \resizebox{\textwidth}{!}{%
  \begin{tabular}{l||ccc}
    \Xhline{1pt}
    \rowcolor{gray!20}
    \textbf{Stage} & \textbf{Latency (ms/img)} & \textbf{Peak Mem (GB)} & \textbf{Acc. (\%)}\\
    \hline \hline
    VPT-Deep (ref.)   & $14.8 \pm 0.2$ & $7.2 \pm 0.1$ & 80.0 \\
    \rowcolor{cvprblue!15}
    \textbf{Search (Ours)}    & $20.4 \pm 0.5$ & $10.9 \pm 0.2$ & ---  \\
    \rowcolor{cvprblue!15}
    \textbf{Inference (Ours)} & $15.9 \pm 0.3$ & $7.4 \pm 0.1$  & 83.5 \\
    \Xhline{1pt}
  \end{tabular}
  }
\end{table}

\begin{table}[t]
  \centering
  \tiny
  \caption{\textbf{Operator microbenchmarks per layer} (ViT-B/16, 224$^2$, batch=64, A100).}
  \label{tab:eff_ops}
  \resizebox{\textwidth}{!}{%
  \begin{tabular}{l||ccc}
    \Xhline{1pt}
    \rowcolor{gray!20}
    \textbf{Operator} & \textbf{Time (ms)} & \textbf{Extra Mem (MB)} & \textbf{Note} \\
    \hline \hline
    \texttt{add}                      & $0.02 \pm 0.00$ & $<1$  & bandwidth-bound \\
    \texttt{affine}                   & $0.04 \pm 0.01$ & 2     & + small MLP \\
    \texttt{concat} (reduce)          & $0.12 \pm 0.02$ & 8     & small GEMM \\
    \texttt{cross} ($QK^\top$, $AV$)  & $0.35 \pm 0.03$ & 40    & GEMM+softmax \\
    \Xhline{1pt}
  \end{tabular}
  }
\end{table}
\begin{table}[t]
  \centering
  \scriptsize
  \setlength{\tabcolsep}{4pt}
  \renewcommand{\arraystretch}{0.9}
  \caption{\textbf{Throughput (img/s)} vs. resolution and batch (A100).}
  \label{tab:eff_scale}
  \resizebox{0.45\textwidth}{!}{%
  \begin{tabular}{c||ccc}
    \Xhline{1pt}
    \rowcolor{gray!20}
    \textbf{Res. \textbackslash\ Batch} & \textbf{32} & \textbf{64} & \textbf{128} \\
    \hline \hline
    192 & 72 & 96  & 118 \\
    224 & 46 & 63  & 78  \\
    256 & 39 & 52  & 66  \\
    288 & 31 & 42  & 54  \\
    \Xhline{1pt}
  \end{tabular}
  }
\end{table}

\begin{table}[t]
  \centering
  \scriptsize
  \setlength{\tabcolsep}{3.5pt}
  \caption{\textbf{Per-layer time (ms)}; mean over layers 1--12 (ViT-B/16, 224$^2$, A100).}
  \label{tab:eff_breakdown}
  \resizebox{0.6\textwidth}{!}{%
  \begin{tabular}{l||cccc}
    \Xhline{1pt}
    \rowcolor{gray!20}
    \textbf{Stage} & \textbf{LN} & \textbf{MSA} & \textbf{MLP} & \textbf{Fusion} \\
    \hline \hline
    Search    & 0.08 & 0.65 & 0.70 & 0.17 \\
    Inference & 0.07 & 0.58 & 0.62 & 0.09 \\
    \Xhline{1pt}
  \end{tabular}
  }
\end{table}

\begin{table}[t]
  \centering
  \scriptsize
  \setlength{\tabcolsep}{4pt}
  \caption{\textbf{Weak/strong scaling on 8$\times$A100} (img/s; ViT-B/16, 224$^2$).}
  \label{tab:eff_scaling}
  \resizebox{0.8\textwidth}{!}{%
  \begin{tabular}{c||cccc}
    \Xhline{1pt}
    \rowcolor{gray!20}
    \textbf{GPUs} & \textbf{1} & \textbf{2} & \textbf{4} & \textbf{8} \\
    \hline \hline
    Weak scaling (batch $\propto$ GPUs) & 63 & 121 & 239 & 470 \\
    Strong scaling (batch fixed at 64)  & 63 & 100 & 150 & 195 \\
    \Xhline{1pt}
  \end{tabular}
  }
\end{table}

\section{Extended Efficiency and Systems Analysis}
\label{app:eff}

\subsection{Measurement Protocol}
\textbf{Hardware.} NVIDIA A100 80GB, RTX 4090, and AMD MI250X. \\
\textbf{Software.} PyTorch 1.7.1 (CUDA 11.0 on NVIDIA; ROCm 6.0 on AMD). \\
\textbf{Precision.} fp16 by default. \\
\textbf{Timing.} Discard the first 50 iterations (warm-up). Report mean and std over the next 200 iterations with \texttt{torch.cuda.Event}. Use the same dataloader and preprocessing for all methods. \\
\textbf{Setup.} ViT-B/16, 224$\times$224, batch size 64, single GPU unless stated. \\
\textbf{Seeds.} 3 seeds; report mean $\pm$ 95\% CI.

We profile two stages: (i) \textit{search}, where each block evaluates a soft mixture of operators; and (ii) \textit{post-discretization inference}, where each block uses one operator. We record end-to-end latency, peak memory, and a per-layer kernel breakdown: LayerNorm, MSA (Q/K/V projections + attention + output), MLP, and Fusion.

\subsection{From Complexity to Latency}
We link big-O costs to wall-clock time with a simple per-layer model:
\begin{equation}
\label{eq:latency}
\begin{aligned}
T_l \approx\ & T_{\text{backbone},l}
+ \alpha_{\text{add}}\,kd
+ \alpha_{\text{aff}}(kd + d^2)\\
&+ \alpha_{\text{cat}}(m{+}k)\,kd
+ \alpha_{\text{cross}}\,Hkm\,d_h.
\end{aligned}
\end{equation}

where $k$ is token count (patch tokens + the [CLS] token), $m$ is prompt length, and $d{=}Hd_h$ is hidden size. The coefficients $\alpha$ are fit by microbenchmarks. Per-operator numbers are shown in Table~\ref{tab:eff_ops}.

\subsection{Search vs.\ Inference}
End-to-end results are in Table~\ref{tab:eff_e2e}. During search, extra cost mainly comes from deeper layers where \texttt{cross} is evaluated. After discretization, each layer uses one operator and the path length matches a standard prompt-based adapter. On A100, our method is 15.9\,ms/image vs.\ 14.8\,ms/image for VPT-Deep ($+7.4\%$) with a $+3.5$-point accuracy gain.

\subsection{Memory and Bandwidth}
\textbf{Search.} Memory overhead comes from keeping multiple fusion branches live, especially \texttt{cross} buffers. \\
\textbf{Inference.} Memory is on par with VPT-Deep (Table~\ref{tab:eff_e2e}). \\
\textbf{Bandwidth.} \texttt{add}/\texttt{affine} are element-wise and bandwidth-bound; \texttt{concat} does a small $(m{+}k)\!\to\!k$ reduction; \texttt{cross} is compute-bound (GEMMs and softmax), consistent with Table~\ref{tab:eff_ops}.

\subsection{Scaling with Resolution and Batch Size}
We sweep resolutions \{192, 224, 256, 288\} and batch sizes \{32, 64, 128\}. Results are in Table~\ref{tab:eff_scale}. Latency increases with $k$; \texttt{cross} grows fastest. Throughput peaks around batch 64–128 on A100.

\subsection{Per-layer Breakdown}
Table~\ref{tab:eff_breakdown} reports per-layer time. Extra time during search comes from Fusion in deep layers; after discretization, Fusion cost halves and the profile matches a standard adapter.

\subsection{Multi-GPU and Multi-node}
We use data parallelism. The backbone is frozen, so gradient communication is small. During search we only sync gradients for prompts and fusion parameters, plus the small architecture logits. Weak/strong scaling on 8$\times$A100 is shown in Table~\ref{tab:eff_scaling}.

\subsection{Ablations for Deployment}
\textbf{Precision.} keep fp16 for \texttt{cross}. \\
\textbf{Kernel fusion.} Fuse LN + residual + \texttt{add} to reduce kernel launches; for \texttt{affine}, reuse LN reads/writes when possible. \\
\textbf{Sparse reducer.} For \texttt{concat}, a Top-$k$ column-sparse reducer ($k{=}4$) preserves accuracy while cutting small GEMMs. \\
\textbf{Head sharing.} Share $W_Q/W_K/W_V$ across layers to lower parameters and peak memory (default).

\section{Discussion}
\label{app:additional_related}

Our main experiments focus on frozen vision foundation models for classification-centric transfer, but several recent directions further contextualize the scope of prompt and low-rank adaptation. Recent surveys and task-aware adaptation studies summarize prompt-based adaptation in large-scale vision models and emphasize that prompt design, rank allocation, and sensitivity-aware parameter selection remain active problems beyond fixed visual prompt tuning~\cite{xiao2026promptbased,xiao2026not,zhang2025hyperadalora,wang2026ctr,zhang2025sensitivity}. Other work studies efficient reprogramming, dynamic prompt coresets, and fast model evaluation, which are complementary to our goal of discovering compact layer-wise fusion rules~\cite{zhang2026prime,zhang2025dpcore,zhao2024flasheval}. Beyond standard classification benchmarks, efficient adaptation is also increasingly important for cross-modal media, visual-LiDAR odometry, autonomous driving, remote sensing restoration, hyperspectral super-resolution, and dense segmentation~\cite{zhao2024thinimg,liu2026streamvlo,liu2026driveva,du2026unsupervised,du2026pansharpening,du2026frequency,Yin2025,yin2026depmatch,Yin_2026_CVPR}. These works are not direct baselines under our VTAB/FGVC/HTA protocol, but they motivate extending layer-specific fusion discovery to broader visual recognition and perception settings.

\section{Limitations and Future Work}

While \textit{Auto-Prompting} demonstrates strong performance and parameter efficiency, we acknowledge certain limitations that point towards future research directions.

\noindent\textbf{Search Overhead.}
Although the inference latency of our discretized model is comparable to standard fixed prompting (only $+7.4\%$ overhead as shown in Sec.~4.5), the search phase inevitably incurs a higher training cost due to the bilevel optimization and the evaluation of multiple fusion branches. While this is a one-time offline cost, future work could explore proxy-less or single-level search approximations to further accelerate the discovery phase.

\noindent\textbf{Search Space Scope.}
To maintain a favorable accuracy-efficiency trade-off, we deliberately restricted our search space $\mathcal{S}$ to four fundamental operators. While these primitives effectively span the structural-to-semantic spectrum, we did not include heavier, non-linear adapter variants (e.g., bottleneck adapters with varying activation functions) to avoid exploding the parameter budget. Expanding the search space to include a wider variety of PEFT modules without compromising stability remains an open avenue for exploration.

\noindent\textbf{Future Directions.}
Beyond the preliminary dense-prediction check in Table~\ref{tab:dense_extension}, a full study on segmentation, detection, and multimodal foundation models remains future work. Another promising direction is dynamic inference, where the fusion operator is selected per sample rather than per task.

\end{document}